# A Multimodal and Explainable Machine Learning Approach to Diagnosing Multi-Class Ejection Fraction from Electrocardiograms

**Author Names**

Catherine Ning[1], Yu Ma[1,2], Cindy Beini Wang[1], Sean McMahon, MD[3]; Joseph Radojevic, MD[3]; Steven Zweibel, MD[3], Dimitris Bertsimas[1]

**Affiliations**

[1] Operations Research Center, Massachusetts Institute of Technology, Cambridge, MA, USA

[2] Wisconsin School of Business, University of Wisconsin–Madison, Madison, WI, USA

[3]Heart and Vascular Institute, Hartford Hospital, Hartford, CT, USA

**Corresponding author**: Dimitris Bertsimas, dbertsim@mit.edu

## Abstract

Left ventricular ejection fraction (LVEF) assessment depends on echocardiography, limiting access in primary care and resource-constrained settings. We developed a multimodal machine-learning framework that combines engineered 12-lead ECG timeseries features with structured EHR variables to classify LVEF into four clinically used strata: normal (≥50%), mildly reduced (40–50%), moderately reduced (30–40%), and severely reduced (<30%). To support model explainability, we identified the most influential ECG and EHR features via SHAP attributions. Using retrospective data from Hartford HealthCare, we trained XGBoost models on 36,784 ECG–echocardiogram pairs from 30,952 outpatients and evaluated temporal generalizability on 19,966 ECGs from a subsequent period. The multimodal model achieved one-vs-rest AUROCs of 0.95 (severe), 0.92 (moderate), 0.82 (mild), and 0.91 (normal), outperforming ECG-only and EHR-only baselines, and maintained performance under temporal validation. This work supports ECG-based, multimodal LVEF stratification as a practical screening and triage aid to prioritize confirmatory imaging where resources are limited.

**Key words:** Electrocardiogram (ECG)-based AI diagnostics, artificial intelligence, Multimodal machine learning, left ventricular ejection fraction (LVEF), heart failure risk stratification, clinical decision support

**Abbreviations:** LVEF – Left ventricular ejection fraction, EHR - electronic health record, HHC - Hartford Healthcare hospital network, AUC/AUROC – Area under the Receiver Operating Characteristic curve, XGBoost – extreme gradient boosting (machine learning model), SHAP – SHapley Additive exPlanations

# Introduction

Left ventricular ejection fraction (LVEF) serves as a fundamental diagnostic and prognostic marker for heart failure and risk stratification for sudden cardiac death [1]. Current clinical practice relies heavily on echocardiography or cardiac magnetic resonance imaging for LVEF assessment—procedures that have limited access and are time-intensive, costly, and require specialized expertise. In contrast, the standard 12-lead electrocardiogram (ECG) is ubiquitous, rapid, and cost-effective, routinely performed across a broad distribution of healthcare environments.

However, LVEF has not historically been evaluated from ECGs, since the subtle features encoding this information are imperceptible to human analysis. This represents a critical opportunity for AI-assisted diagnostic tools that can uncover these latent signals. By extracting this hidden information within routine ECGs, we could provide actionable recommendations, such as automated alerts suggesting further patient evaluation (i.e. an echocardiogram) when ECG patterns indicate a risk of reduced LVEF. The ability to derive previously inaccessible insights from a low-cost, widely available test is transformative, with implications for both high-income settings and low- and middle-income countries where access remains limited [2].

Existing ECG-based AI models have largely focused on binary classification of systolic function, typically distinguishing between "normal" and "reduced" ejection fraction using a single cutoff (e.g., <35% in [3]). Other recent studies have shifted toward predicting long-term adverse events [4], diastolic dysfunction at preserved EF levels [5], or longitudinal changes in heart failure patients [6]. While clinically useful, these frameworks often prioritize binary or risk-based outcomes and may not capture the granular distinctions between mild, moderate, and severe systolic dysfunction—categories that directly inform prognosis and therapeutic decision-making. In addition, multimodal learning strategies that integrate complementary data sources remain less explored in this setting. Many published approaches rely solely on ECG tracings [3,5] or EHR-derived features [6], rather than jointly modeling them. Moreover, existing high-performing approaches typically employ deep learning architectures, which can impose computational and operational barriers to deployment in routine clinical workflows [7]. Our approach instead unifies ECG time-series–derived structured features and EHR data within an XGBoost-based tabular learning framework, aligning with the Holistic AI in Medicine paradigm that emphasizes gains from integrating heterogeneous modalities [8].

In this work, we (i) develop a multimodal framework that transforms 12-lead ECG time series into structured features that can be readily combined with tabular EHR data, (ii) formulate and evaluate a four-class LVEF stratification model using clinically meaningful thresholds (<30%, 30–40%, 40–50%, ≥50%) to extend beyond prior binary formulations, and (iii) provide feature-level attribution analyses to support explainability of which ECG- and EHR-derived variables are most associated with model predictions, while prioritizing a computationally efficient pipeline designed for clinical deployment.

# Results

## Development Cohort Performance

On the internal held-out test set, the multimodal XGBoost classifier integrating engineered ECG time-series features with EHR variables achieved the strongest discrimination across all four LVEF categories (Table 1). One-versus-rest AUROCs were 0.95 (95% CI 0.93–0.96) for severe LVEF reduction (<30%), 0.92 (0.89–0.94) for moderate reduction (30–40%), 0.82 (0.80–0.84) for mild reduction (40–50%), and 0.91 (0.89–0.92) for normal LVEF (≥50%).

Unimodal baselines showed consistent but weaker performance. ECG-only discrimination was strong, particularly for severe reduction (AUROC 0.94, 0.92–0.96), whereas EHR-only performance was lower overall (e.g., 0.87 [0.81–0.92] for severe reduction and 0.76 [0.72–0.80] for mild reduction). Multimodal fusion yielded incremental improvements beyond ECG-only inputs for each category, with the largest absolute gain observed for the moderate class (0.88 to 0.92 AUROC), suggesting complementary information captured by routine EHR variables (Table 1).

Given substantial class imbalance in the development cohort (normal LVEF representing most observations and severe reduction comprising a small minority), threshold-dependent metrics were more variable across categories. At the per-class operating point that maximized one-versus-rest F1, minority-class F1 scores were modest (severe: 0.35 [0.32–0.40]; moderate: 0.27 [0.25–0.32]; mild: 0.27 [0.25–0.29]), while sensitivity ranged from 0.57–0.73 and specificity from 0.82–0.95 for the reduced LVEF categories (Table 2). In contrast, the normal class achieved high F1 (0.95 [0.95–0.96]) and very high sensitivity (0.99 [0.98–0.99]); however, its one-versus-rest specificity was low (0.26 [0.20–0.33]) because optimizing F1 for the dominant class can yield a permissive threshold that inflates predicted normal assignments. For this reason, AUROC is emphasized as the primary summary of ranking performance across classes, while Table 2 provides an operating-point view for context.

Receiver operating characteristic curves for the multimodal model on the internal held-out test set are shown in Figure 1. The AUROC values displayed in the legend reflect a single evaluation on the full test set and may differ slightly from the bootstrapped averages reported in Table 1.

**External temporal validation**

Temporal external validation on a subsequent cohort showed that discrimination remained stable over time and that multimodal integration continued to provide consistent gains over unimodal models (Table 3). On the temporal cohort, multimodal AUROCs were 0.93 (0.92–0.94) for severe reduction, 0.89 (0.88–0.90) for moderate reduction, 0.81 (0.80–0.81) for mild reduction, and 0.89 (0.88–0.89) for normal LVEF. Compared to internal evaluation, the largest AUROC decrease was observed for the moderate class (0.92 to 0.89), while changes for other classes were smaller in magnitude (Table 1 vs. Table 3).

At the F1-maximizing operating point in the temporal cohort, F1 scores for reduced LVEF classes remained modest (severe: 0.31 [0.28–0.32]; moderate: 0.29 [0.28–0.30]; mild: 0.27 [0.26–0.28]), with sensitivities of 0.56–0.60 and specificities of 0.83–0.95 (Table 4). Normal one-vs-rest specificity remained low at the F1-maximizing threshold for the same prevalence-driven reason noted above. Temporal ROC curves are provided in Figure 2.

**Feature attribution and model explainability**

Class-specific SHAP analyses identified a small and consistent subset of features driving model outputs across the four LVEF categories (Figure 3, Supplemental Figures S2-S4). High-impact features spanned both modalities and included ECG-derived voltage and morphology summaries (e.g., Lead I and V6 QR-interval amplitude statistics, lead-specific amplitude-band descriptors, and frequency-domain complexity measures such as spectral entropy), as well as clinically plausible EHR variables (e.g., past cardiomyopathy diagnosis, systolic blood pressure, and sex). Across classes, the SHAP beeswarm plots further illustrate how feature values (high vs. low) align with positive or negative contributions to the corresponding class output. We provide an expanded narrative summary of the SHAP drivers and directionality for Figure 3 in the Supplemental Note 1.

Feature ranking stability was high under bootstrap resampling of the test set. Across 20 bootstrap resamples, the top 10 feature sets had a high overlap, indicating that the dominant explanatory features were robust to sampling variability (Supplemental Figure S5). The most stable features appeared in 95–100% of bootstrap resamples, including Lead I QR-interval average/median amplitude, ICD-10 cardiomyopathy and ischemic cardiomyopathy groupings, sex, Lead V6 QR-interval median amplitude, and selected time-series complexity descriptors.

Dependence plots for the top ECG-derived features (Supplemental Figure S6) showed an approximately monotonic relationship between feature magnitude and SHAP contribution, with higher Lead I QR-interval amplitude summary statistics associated with more negative SHAP values (i.e., reduced model output for the plotted class), consistent with a directionally coherent effect in this cohort. Lastly, a list of the top 10 features with highest mean absolute SHAP values is provided for each outcome class in Supplemental Figure S7).

# Discussion

This study demonstrates that left ventricular ejection fraction can be estimated from standard ECGs, reinforcing and extending prior evidence that clinically meaningful information can be extracted from signals not traditionally used for this purpose. By incorporating routinely collected EHR data, our multimodal framework improves discrimination relative to unimodal ECG-only and EHR-only baselines using inputs that are widely available across healthcare systems.

An additional contribution of this work is model explainability at the feature level. Rather than treating "explainability" as an intrinsic property of the learning algorithm, we use post-hoc feature attribution (SHAP) to identify which engineered ECG and EHR features contribute most strongly to the model outputs. Because the ECG representation is constructed from clinically recognizable waveform components and established statistical transforms of time series, these attributions can be mapped back to interpretable signal characteristics (e.g., lead-specific morphology, interval statistics, frequency-domain descriptors), supporting clinical and engineering review.

Compared with prior deep learning–based approaches, our results show that a tabular learning pipeline can achieve competitive discrimination while offering practical advantages in efficiency and deployment: (1) the tsfresh and ECG feature extraction software is open-source and CPU-compatible, facilitating integration into hospital systems; (2) XGBoost models typically require substantially fewer computational resources than deep neural networks [7]; and (3) our preprocessing relies on standard techniques with minimal implementation barriers. Importantly, our four-class classification scheme adds clinically relevant granularity beyond binary models, supporting more nuanced risk stratification. Despite the increased difficulty of the modeling problem under this higher granularity, we obtain strong out-of-sample one-versus-rest AUROCs that are comparable to existing work: to contextualize against the 50% [6] and 35% [3] cutoffs used in prior publications, we report an out-of-sample AUC of 0.91 for a binary threshold at 50% LVEF, and a bracketing performance of 0.92–0.93 AUC at the adjacent thresholds 30% and 40%[1], albeit we acknowledge that the datasets are different from those used in those papers.

These findings contribute to the growing literature on AI for cardiovascular screening while providing complementary functionality to several recent directions. The ECG-MACE model [4] targets broad long-term adverse events and multi-task risk stratification rather than immediate diagnostic classification of systolic function. Lee et al. [5] focuses on elevated filling pressures and HFpEF-related physiology that can be partially independent of EF values. Longitudinal EHR-based models [6] use temporal data to predict LVEF changes, whereas our approach uses a single-point 12-lead ECG (augmented by routinely available EHR variables) to provide instantaneous four-class stratification aligned with physician-approved EF thresholds. In this sense, the proposed framework complements—rather than duplicates—existing risk-stratification tools.

This study has several limitations. First, the development and temporal validation cohorts were restricted to outpatients from Hartford HealthCare, reflecting our goal of a screening tool suitable for primary care. This focus enhances relevance to that clinical context but may limit generalizability to other health systems, geographic regions, or inpatient and emergency populations. Second, both cohorts exhibit substantial class imbalance, with severe LVEF reduction representing a small fraction of cases. While AUROC indicates strong overall discrimination, threshold-dependent metrics (e.g., F1) are more sensitive to class prevalence and can be lower for minority categories in real-world clinical populations. Third, although we report temporal validation demonstrating stability across a subsequent period, prospective evaluation within clinical workflows is required to characterize real-world impact, including how model outputs would influence referral patterns and downstream testing.

[1] Because our EPIC-derived labels are available at 30% and 40% thresholds (not 35%), we report at these adjacent cutoffs, which provides contextual, though not direct, comparability.

Consistent with a screening and prioritization application, we emphasize discriminatory performance for identifying patients at higher risk of reduced LVEF rather than interpreting predicted probabilities as calibrated estimates of absolute risk across classes. Before clinical use in probability-based decision pathways, prospective evaluation is needed to assess workflow integration, monitoring burden, and downstream utilization of echocardiography, and to establish appropriate site-specific thresholding and recalibration strategies.

From a clinical perspective, the model is best viewed as a decision-support aid that could surface patients warranting confirmatory imaging when echocardiography access is limited. In primary care, it could support targeted referral for echocardiography among patients with nonspecific symptoms or elevated risk. In longitudinal specialty settings (e.g., heart failure or cardio-oncology clinics), it may support earlier detection of clinically meaningful EF deterioration between scheduled imaging intervals. More broadly, ECG-based screening could reduce unnecessary echocardiograms in low-risk patients while helping identify patients with substantial EF reduction who might otherwise go unrecognized, enabling timely referral and care escalation.

Overall, this work supports a pragmatic multimodal approach for four-class LVEF classification from widely available data sources. By combining engineered ECG time-series features with routine EHR variables in a computationally efficient pipeline and providing feature-level attributions via SHAP for model explainability, we offer a deployable framework that complements existing ECG-AI literature and can help prioritize confirmatory imaging in settings where resources are constrained.

# Methods

## Study design and cohorts

We conducted a retrospective study using data from Hartford HealthCare (HHC), Connecticut's largest healthcare network serving both rural and urban patient populations. We included adult outpatients who received both a standard 12-lead ECG and an echocardiogram with physician-reviewed LVEF assessment between March 2020 and November 2023. ECG–echocardiogram pairs were eligible if the ECG was acquired within 14 days of the echocardiogram, a window selected to balance temporal stability of cardiac function with adequate sample size. When multiple ECGs fell within the eligible window for a given echocardiogram, we retained the temporally closest ECG.

LVEF was obtained from the echocardiography reporting database and mapped into four outcome categories: normal (≥50%), mildly reduced (40–50%), moderately reduced (30–40%), and severely reduced (<30%). Echocardiograms flagged as "poor quality", or "artifact" were excluded. The final development cohort contained 36,784 ECG examinations from 30,952 unique patients, with class prevalences of 88.35% normal, 5.93% mildly reduced, 3.47% moderately reduced, and 2.26% severely reduced. A flowchart of the study cohort can be found in Supplemental Figure S1.

**ECG data and time-series feature engineering**

Raw ECG signals consisted of 10-s recordings sampled at 500 Hz from leads I, II, and V1–V6. Leads III, aVR, aVL, and aVF were derived from leads I and II using Einthoven's law to form a complete 12-lead representation.

Feature generation combined three complementary components:

1. **Signal preprocessing.** We applied a 0.5-Hz high-pass Butterworth filter (order 5), removed powerline interference, and z-score standardized each lead signal to reduce baseline drift and improve comparability across recordings.
2. **Clinically motivated ECG features.** We extracted 127 ECG features capturing RR-interval statistics (e.g., mean, standard deviation, minimum, maximum), heart-rate metrics, inter-beat interval measures, and PQRST morphology measurements using established ECG feature sets. These features were intended to preserve clinically recognizable signal characteristics while yielding a structured representation suitable for tabular learning [9].
3. **Automated time-series features.** We used the `tsfresh` library to compute 777 time-series descriptors spanning statistical moments, stationarity- and autocorrelation-based features, information-theoretic measures, model-based summaries, and frequency-domain characteristics [10]. This step was included to capture complex waveform properties beyond canonical ECG measurements.

The ECG feature set was formed by concatenating the clinical and tsfresh features across leads after preprocessing.

**Electronic health record features**

We augmented ECG features with structured EHR variables extracted from the six months preceding the echocardiogram, including demographics, body measurements, medication history, and comorbidities represented by ICD-10 diagnosis groupings. To reduce sparsity and improve robustness, we retained the 50 most frequent diagnosis groupings and 50 most frequent medication groupings. To mitigate label leakage, we excluded LVEF-proximal clinical variables (e.g., heart failure ICD-10 I50 and other explicitly LVEF-defining codes). Categorical variables were one-hot encoded.

For cohort-level descriptive statistics, p values were computed using $\chi^2$ tests for binary/categorical variables and Kruskal–Wallis tests for continuous variables. Summary statistics for EHR features are provided in Supplementary Tables 1–2.

**Model development and evaluation**

We trained supervised classifiers to predict the four LVEF categories from (i) EHR-only features, (ii) ECG-only features, and (iii) the combined multimodal feature set. The development cohort was split into training (80%), validation (10%), and held-out test (10%) partitions using stratification by LVEF class; patient-level splitting was enforced to prevent the same patient contributing ECGs to multiple partitions. We evaluated multiple model families (logistic regression, random forest, and gradient boosting), and selected XGBoost as the primary model due to consistently superior validation performance and favorable computational efficiency.

Primary performance was assessed on the held-out test set using one-versus-rest AUROC for each LVEF category. Uncertainty intervals for AUROC were estimated via non-parametric bootstrapping of the held-out test set and are reported as bootstrapped 95% confidence intervals. For threshold-dependent metrics (F1, sensitivity, specificity), we selected class-specific decision thresholds that maximize per-class one-versus-rest F1 on the validation set and then evaluated these thresholds on the corresponding test set. All analyses were implemented in Python; runtime measurements are reported for transparency (training without hyperparameter search on a single NVIDIA L40s GPU was approximately 3.33 minutes; ten bootstrap resamples required additional 30 seconds).

**Temporal external validation**

To assess temporal stability and generalizability, we performed temporal external validation [11] using a subsequent outpatient dataset from January 2024 through July 2025. This dataset contained 22,287 ECG examinations paired with the temporally closest echocardiogram LVEF label. Applying the same inclusion criteria and preprocessing/feature extraction pipeline (including the 14-day ECG–echocardiogram pairing window), we obtained 19,966 ECG examinations from 18,057 unique patients with accompanying EHR features. Class prevalences in the temporal cohort were 88.56% normal, 6.01% mildly reduced, 3.34% moderately reduced, and 2.10% severely reduced. Model training was performed on the original development cohort only, and all temporal validation results were computed on this held-out future cohort.

**Model explainability via SHAP**

To characterize feature-level drivers of the XGBoost predictions, we used SHapley Additive exPlanations (SHAP) [12], a game-theoretic attribution method for estimating each feature's contribution to a model output. SHAP values were computed on the held-out test set so that attributions reflect model behavior on unseen data. For the four-class classifier, we computed class-specific SHAP values for each class output (default parameter "margin") and summarized global importance using the mean absolute SHAP value across samples; class-conditional summaries are visualized using SHAP beeswarm plots (Figure 3, Supplemental Figures S2-S4).

Because engineered feature identifiers (particularly time-series descriptors) are not inherently human-readable, we applied a post-hoc label mapping to translate raw feature names into standardized, clinician-interpretable descriptors (e.g., lead, waveform component, and feature family). This mapping was used for visualization only; model inputs and computed SHAP values were generated using the original feature matrix. A complete mapping table linking display labels to raw feature definitions is provided as supplementary material.

To assess the stability of global feature rankings, we performed a bootstrap stability analysis on the test set. We generated $B = 20$ bootstrap resamples (with replacement) of the test set, recomputed mean absolute SHAP values for each resample, and extracted the top-$k = 10$ features per resample. Ranking stability was quantified using pairwise Jaccard similarity between top 10 feature sets across bootstrap resamples, and feature-frequency (the proportion of bootstrap resamples in which a feature appeared in the top 10) was used to identify consistently selected features. To visualize the functional relationship between high-impact ECG-derived features and their contribution to the model output, we generated dependence plots for top 2

ECG-derived features and their average SHAP values across all 4 classes. We overlaid a fitted linear trend to summarize directionality, while using the full scatter distribution to reveal potential non-linear patterns.

# Data Availability

This study received Institutional Review Board approval from HHC for the use of retrospective electronic health record data. All patient data management adhered to HIPAA regulations and cannot be shared publicly due to privacy restrictions.

# Code Availability

De-identified analysis code and feature extraction pipelines are available upon reasonable request to the corresponding author.

# Human Ethics and Consent to Participate declarations

Not applicable.

# Acknowledgements

This research was supported by internal funding from Hartford HealthCare's Center for AI Innovation and academic grants in healthcare awarded to Prof. Dimitris Bertsimas.

# Author Contributions

CN, YM, SM, SZ, and DB conceived the study and designed the experiments. CN and YM performed the data collection and clinical validation. CN and YM developed the machine learning models and analyzed the data. CN and CW wrote the initial manuscript. CN, YM, CW, SM, JR, SZ, and DB contributed to the interpretation of the results and provided critical feedback on the manuscript. CN, YM, CW, SM, JR, SZ, and DB approved the final version of the manuscript.

## Competing Interests

All authors declare no financial or non-financial competing interests.

## References

1. Cook, D. A., Oh, S. Y. & Pusic, M. V. Accuracy of physicians' electrocardiogram interpretations: A systematic review and meta-analysis. *JAMA Intern. Med.* **180**, 1461–1471 (2020).
2. World Health Organization. Cardiovascular diseases (CVDs) fact sheet. https://www.who.int/news-room/fact-sheets/detail/cardiovascular-diseases-(cvds).
3. Attia, Z. I. *et al.* Screening for cardiac contractile dysfunction using an artificial intelligence-enabled electrocardiogram. *Nat. Med.* **25**, 70–74 (2019).
4. Kalmady, S. V. *et al.* A multitask deep learning model utilizing electrocardiograms for major cardiovascular adverse events prediction. *npj Digit. Med.* **8**, 24 (2025).
5. Lee, H. J. *et al.* Artificial intelligence-enabled ECG for left ventricular diastolic function and filling pressure. *npj Digit. Med.* **7**, 133 (2024).
6. Kim, M. *et al.* Prediction of left ventricular ejection fraction changes in heart failure patients using machine learning and electronic health records: A multi-site study. *npj Digit. Med.* **6**, 149 (2023).

7. Han, Y. *et al.* AI for regulatory affairs: Balancing accuracy, interpretability, and computational cost in medical device classification. *arXiv* 2505.18695 (2025).
8. Soenksen, L. R. *et al.* Integrated multimodal artificial intelligence framework for healthcare applications. *npj Digit. Med.* **5**, 149 (2022).
9. Bertsimas, D. *et al.* Machine learning for real-time heart disease prediction. *IEEE J. Biomed. Health Inform.* **25**, 3627–3637 (2021).
10. Christ, M. *et al.* Time series feature extraction on basis of scalable hypothesis tests (tsfresh – A Python package). *Neurocomputing* **307**, 72–77 (2018).
11. de Hond, A. A. H. *et al.* Perspectives on validation of clinical predictive algorithms. *npj Digit. Med.* **6**, 86 (2023).
12. Lundberg, S. M. & Lee, S.-I. A unified approach to interpreting model predictions. *Adv. Neural Inf. Process. Syst.* **30**, 4765–4774 (2017).

# Tables & Figures

**Table 1:** AUC performance of our XGBoost classifier on the internal hold-out test set, with bootstrapped 95% confidence intervals in brackets, demonstrating the benefit of including multiple data modalities

| LVEF Outcome Class | EHR Only | ECG Only | Multimodal (EHR+ECG) |
|---|---|---|---|
| Severe (<30%) | 0.87 [0.81-0.92] | 0.94 [0.92-0.96] | **0.95 [0.93-0.96]** |
| Moderate (30-40%) | 0.86 [0.82-0.90] | 0.88 [0.86-0.91] | **0.92 [0.89-0.94]** |
| Mild (40-50%) | 0.76 [0.72-0.80] | 0.80 [0.78-0.83] | **0.82 [0.80, 0.84]** |
| Normal (≥50%) | 0.84 [0.82-0.86] | 0.89 [0.87-0.90] | **0.91 [0.89-0.92]** |

**Table 2:** F1, sensitivity and specificity on internal hold-out test set of our XGBoost multimodal classifier. Sensitivity/specificity are computed in a one-vs-rest manner for each class, with thresholds selected to maximize per-class F1. For the Normal class (dominant prevalence), maximizing F1 can yield a permissive threshold that inflates predicted Normal, reducing one-vs-rest specificity; we therefore interpret AUROC as the primary summary of ranking performance across classes.

| LVEF Class | F1 (95% CI) | Sensitivity | Specificity |
|---|---|---|---|
| Severe (<30%) | 0.35 [0.32, 0.40] | 0.73 [0.67, 0.80] | 0.95 [0.94, 0.96] |
| Moderate (30-40%) | 0.27 [0.25, 0.32] | 0.57 [0.54, 0.59] | 0.91 [0.90, 0.91] |
| Mild (40-50%) | 0.27 [0.25, 0.29] | 0.63 [0.59, 0.68] | 0.82 [0.81, 0.83] |
| Normal (≥50%) | 0.95 [0.95-0.96] | 0.99 [0.98, 0.99] | 0.26 [0.20, 0.33] |

**Table 3:** Temporal external validation performance of our XGBoost classifier.

| LVEF Class | EHR Only | ECG Only | Multimodal |
|---|---|---|---|
| Severe (<30%) | 0.84 [0.83,0.85] | 0.92 [0.91,0.92] | **0.93 [0.92,0.94]** |
| Moderate (30-40%) | 0.79 [0.77,0.80] | 0.86 [0.85,0.87] | **0.89 [0.88, 0.90]** |
| Mild (40-50%) | 0.74 [0.73,0.75] | 0.78 [0.77,0.78] | **0.81 [0.80, 0.81]** |
| Normal (≥50%) | 0.80 [0.79,0.80] | 0.86 [0.85,0.86] | **0.89 [0.88, 0.89]** |

**Table 4**: F1, sensitivity and specificity on temporal external validation cohort of our XGBoost multimodal classifier. Sensitivity and specificity/recall are provided at the operating point maximizing the F1 score.

| LVEF Class | F1 score | Sensitivity | Specificity |
|---|---|---|---|
| Severe (<30%) | 0.31 [0.28, 0.32] | 0.58 [0.55, 0.61] | 0.95 [0.95, 0.96] |
| Moderate (30-40%) | 0.29 [0.28, 0.30] | 0.60 [0.58, 0.63] | 0.91 [0.91, 0.91] |
| Mild (40-50%) | 0.27 [0.26, 0.28] | 0.56 [0.54, 0.59] | 0.83 [0.83, 0.84] |
| Normal (≥50%) | 0.95 [0.94, 0.95] | 0.98 [0.97, 0.99] | 0.29 [0.23, 0.35] |

**Figure 1:** ROC-AUC curves of the four-class multimodal XGBoost model evaluated on the internal held-out test set. Note that the AUC scores in the legend differ slightly from the bootstrapped averages in Tables 1 and 3 since they have been generated using one full prediction on the test sets.

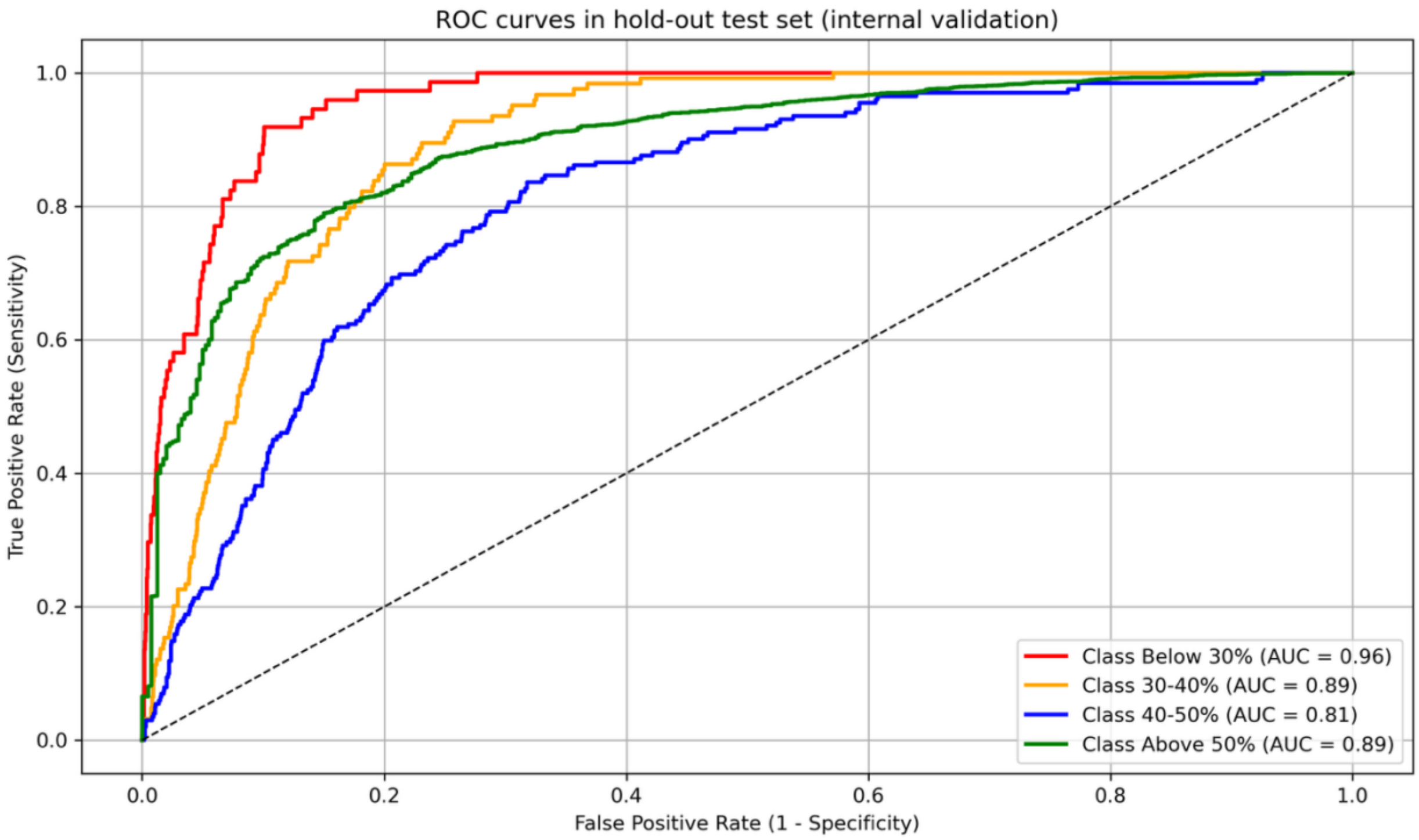

**Figure 2:** ROC-AUC curves of the four-class multimodal XGBoost model evaluated on the temporal external validation cohort.

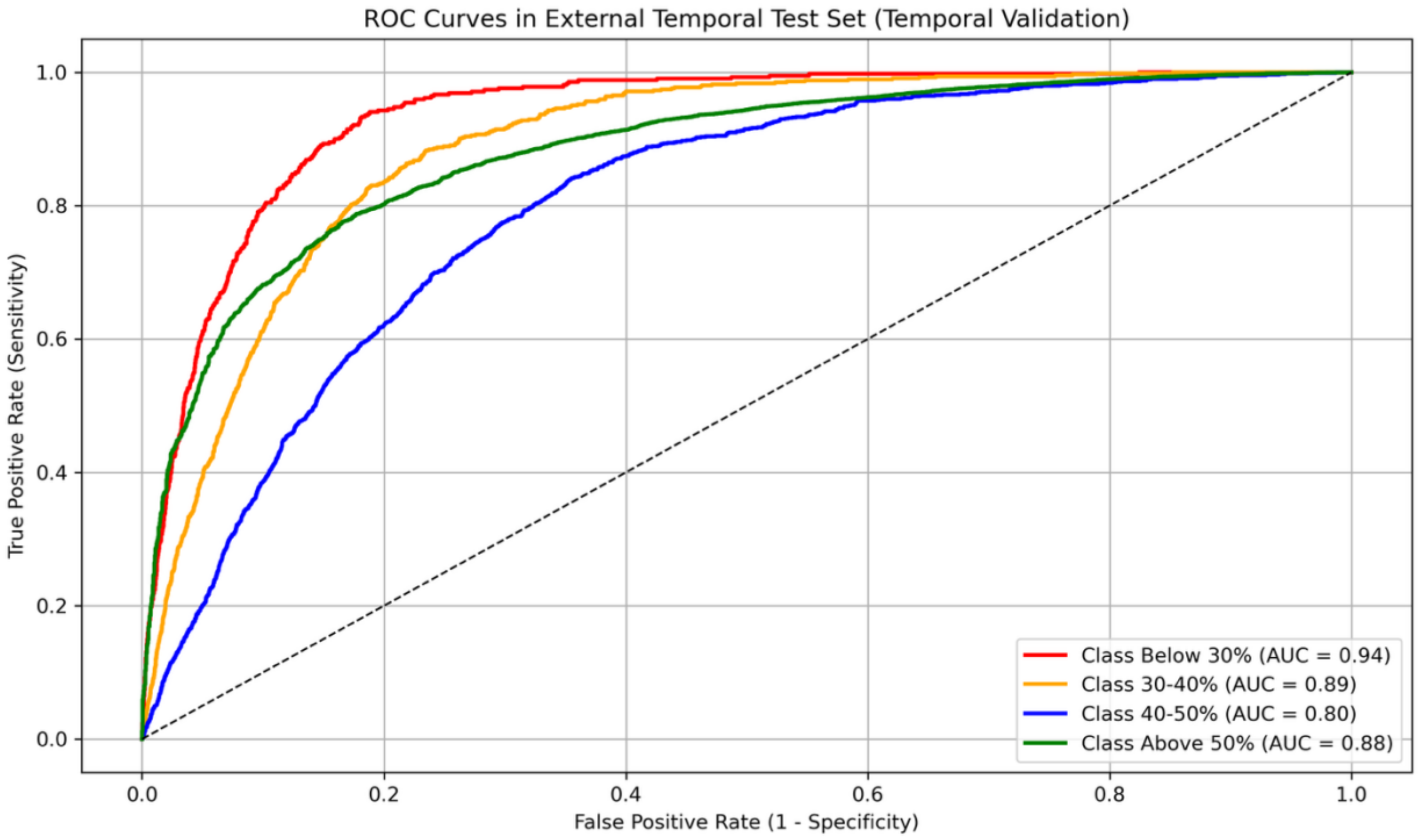

**Figure 3:** Class-specific SHAP analysis of our multimodal XGBoost classifier for class "normal LVEF" (≥50%). Positive SHAP values increase the model's score for the "normal LVEF" class; negative values decrease the class score. Details on how to read this plot and deeper results interpretation are in Supplemental Note 1.

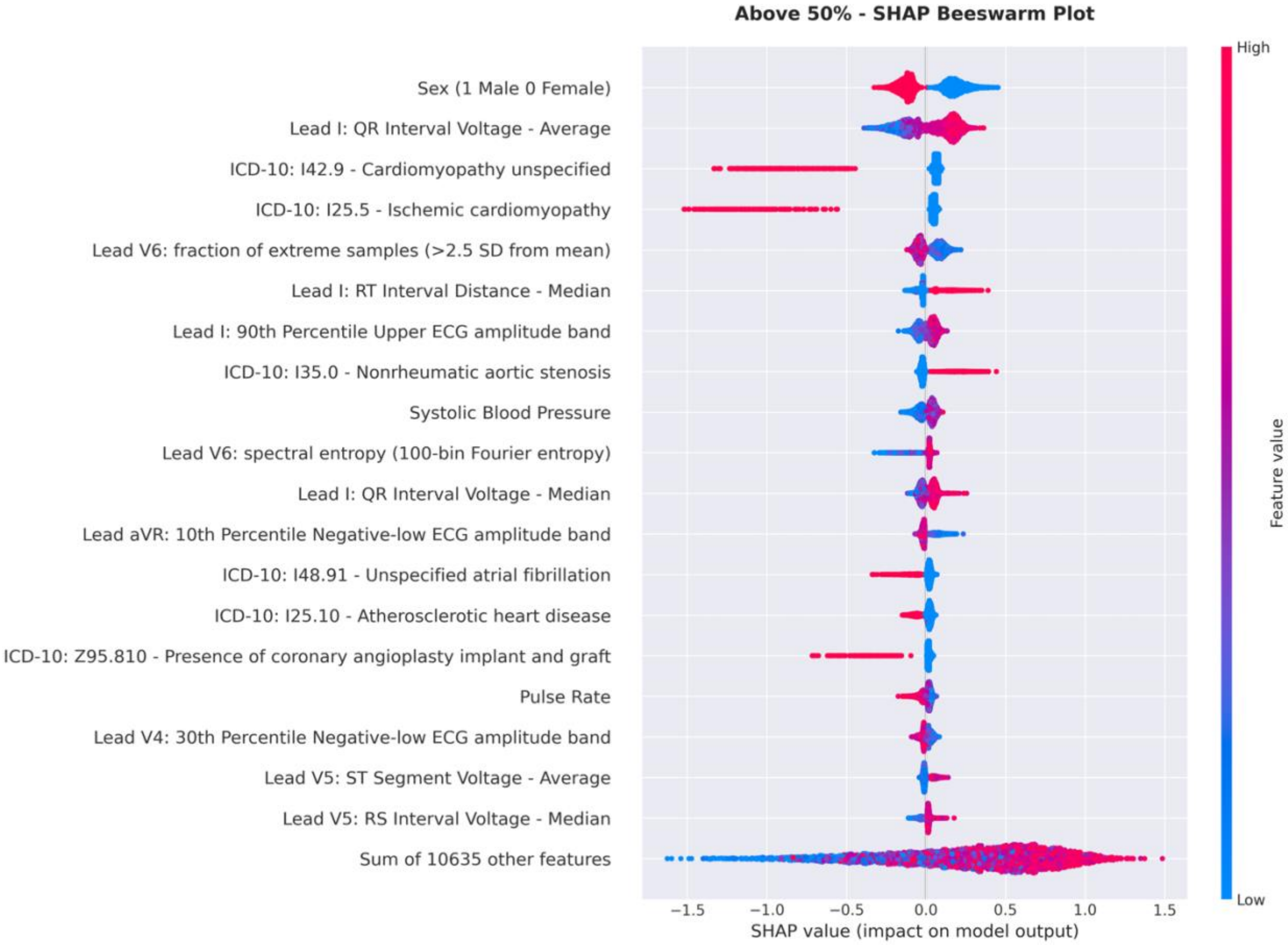

# Supplemental Material

**Supplemental Figure S1. Flow chart of study cohort.**

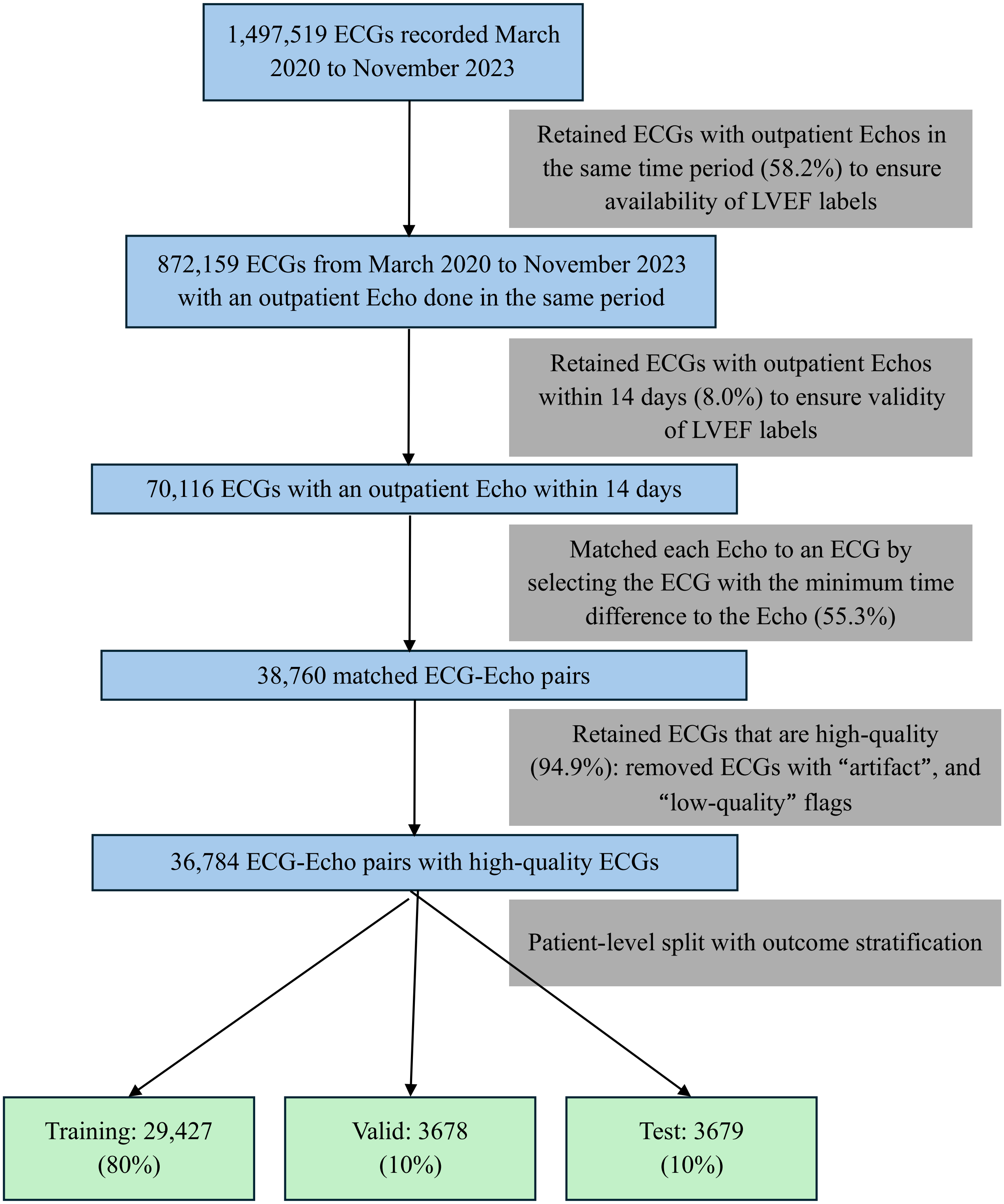

**Supplemental Table 1: Development cohort summary statistics of electronic health record data.** These were used as predictor features in our model training, starting with the 50 most frequent ICD-10 code groups, followed by the 50 most frequent medication types, continuous variables and one-hot-encoded categorical variables on body measurements and demographics. A full list of the predictor features extracted from the ECG data is not provided as it would be too large to be provided in a summary statistics table. P-values for binary and categorical variables were calculated with the Chi-squared test, for continuous variables with the Kruskal-Wallis test.

| Characteristic | Overall | Severely Reduced LVEF | Moderately Reduced LVEF | Mildly Reduced LVEF | Normal LVEF | P-Value |
|---|---|---|---|---|---|---|
| **C50** | 653 (1.8%) | 9 (1.1%) | 14 (1.1%) | 31 (1.4%) | 599 (1.8%) | 0.045 |
| **D50** | 1099 (3.0%) | 30 (3.6%) | 53 (4.2%) | 75 (3.4%) | 941 (2.9%) | 0.022 |
| **D64** | 1724 (4.7%) | 58 (7.0%) | 85 (6.7%) | 109 (5.0%) | 1472 (4.5%) | <0.001 |
| **E03** | 2129 (5.8%) | 49 (5.9%) | 83 (6.5%) | 117 (5.4%) | 1880 (5.8%) | 0.587 |
| **E11** | 4730 (12.9%) | 177 (21.3%) | 234 (18.3%) | 384 (17.6%) | 3935 (12.1%) | <0.001 |
| **E55** | 2537 (6.9%) | 45 (5.4%) | 57 (4.5%) | 109 (5.0%) | 2326 (7.2%) | <0.001 |
| **E66** | 3538 (9.6%) | 72 (8.7%) | 101 (7.9%) | 168 (7.7%) | 3197 (9.8%) | 0.001 |
| **E78** | 15555 (42.3%) | 409 (49.2%) | 614 (48.1%) | 1047 (48.0%) | 13485 (41.5%) | <0.001 |
| **E87** | 1562 (4.2%) | 78 (9.4%) | 90 (7.1%) | 117 (5.4%) | 1277 (3.9%) | <0.001 |
| **F41** | 3055 (8.3%) | 63 (7.6%) | 87 (6.8%) | 147 (6.7%) | 2758 (8.5%) | 0.005 |
| **G47** | 3166 (8.6%) | 87 (10.5%) | 145 (11.4%) | 219 (10.0%) | 2715 (8.4%) | <0.001 |
| **G89** | 2094 (5.7%) | 36 (4.3%) | 73 (5.7%) | 129 (5.9%) | 1856 (5.7%) | 0.377 |
| **I10** | 18573 (50.5%) | 479 (57.6%) | 748 (58.6%) | 1273 (58.4%) | 16073 (49.5%) | <0.001 |
| **I21** | 1063 (2.9%) | 73 (8.8%) | 114 (8.9%) | 137 (6.3%) | 739 (2.3%) | <0.001 |
| **I25** | 7468 (20.3%) | 360 (43.3%) | 600 (47.0%) | 854 (39.2%) | 5654 (17.4%) | <0.001 |
| **I34** | 2576 (7.0%) | 109 (13.1%) | 158 (12.4%) | 236 (10.8%) | 2073 (6.4%) | <0.001 |
| **I35** | 4019 (10.9%) | 86 (10.3%) | 148 (11.6%) | 226 (10.4%) | 3559 (11.0%) | 0.659 |
| **I47** | 1921 (5.2%) | 153 (18.4%) | 162 (12.7%) | 204 (9.4%) | 1402 (4.3%) | <0.001 |
| **I48** | 8742 (23.8%) | 368 (44.3%) | 570 (44.7%) | 949 (43.5%) | 6855 (21.1%) | <0.001 |
| **I49** | 2415 (6.6%) | 91 (11.0%) | 135 (10.6%) | 236 (10.8%) | 1953 (6.0%) | <0.001 |
| **J44** | 1326 (3.6%) | 43 (5.2%) | 80 (6.3%) | 122 (5.6%) | 1081 (3.3%) | <0.001 |
| **J45** | 1621 (4.4%) | 39 (4.7%) | 45 (3.5%) | 82 (3.8%) | 1455 (4.5%) | 0.168 |
| **K21** | 4105 (11.2%) | 83 (10.0%) | 129 (10.1%) | 208 (9.5%) | 3685 (11.3%) | 0.025 |
| **M25** | 2323 (6.3%) | 30 (3.6%) | 76 (6.0%) | 123 (5.6%) | 2094 (6.4%) | 0.004 |
| **M54** | 2843 (7.7%) | 47 (5.7%) | 92 (7.2%) | 155 (7.1%) | 2549 (7.8%) | 0.064 |
| **M79** | 2037 (5.5%) | 44 (5.3%) | 54 (4.2%) | 120 (5.5%) | 1819 (5.6%) | 0.214 |
| **N18** | 2494 (6.8%) | 136 (16.4%) | 184 (14.4%) | 227 (10.4%) | 1947 (6.0%) | <0.001 |
| **N40** | 1694 (4.6%) | 60 (7.2%) | 106 (8.3%) | 166 (7.6%) | 1362 (4.2%) | <0.001 |
| **R00** | 4814 (13.1%) | 75 (9.0%) | 129 (10.1%) | 199 (9.1%) | 4411 (13.6%) | <0.001 |
| **R06** | 7000 (19.0%) | 247 (29.7%) | 304 (23.8%) | 452 (20.7%) | 5997 (18.5%) | <0.001 |
| **R07** | 5150 (14.0%) | 85 (10.2%) | 130 (10.2%) | 220 (10.1%) | 4715 (14.5%) | <0.001 |
| **R10** | 2212 (6.0%) | 34 (4.1%) | 71 (5.6%) | 98 (4.5%) | 2009 (6.2%) | 0.001 |
| **R42** | 2161 (5.9%) | 41 (4.9%) | 63 (4.9%) | 95 (4.4%) | 1962 (6.0%) | 0.003 |
| **R53** | 2224 (6.0%) | 56 (6.7%) | 87 (6.8%) | 131 (6.0%) | 1950 (6.0%) | 0.539 |
| **R60** | 2191 (6.0%) | 68 (8.2%) | 106 (8.3%) | 162 (7.4%) | 1855 (5.7%) | <0.001 |
| **R73** | 2233 (6.1%) | 47 (5.7%) | 72 (5.6%) | 110 (5.0%) | 2004 (6.2%) | 0.158 |
| **R94** | 3650 (9.9%) | 78 (9.4%) | 112 (8.8%) | 196 (9.0%) | 3264 (10.0%) | 0.188 |
| **Z00** | 4094 (11.1%) | 44 (5.3%) | 83 (6.5%) | 182 (8.3%) | 3785 (11.6%) | <0.001 |
| **Z01** | 4607 (12.5%) | 137 (16.5%) | 173 (13.6%) | 292 (13.4%) | 4005 (12.3%) | 0.001 |
| **Z11** | 3009 (8.2%) | 94 (11.3%) | 123 (9.6%) | 188 (8.6%) | 2604 (8.0%) | 0.001 |
| **Z12** | 4155 (11.3%) | 54 (6.5%) | 108 (8.5%) | 196 (9.0%) | 3797 (11.7%) | <0.001 |
| **Z17** | 356 (1.0%) | 6 (0.7%) | 7 (0.5%) | 17 (0.8%) | 326 (1.0%) | 0.252 |
| **Z20** | 2416 (6.6%) | 67 (8.1%) | 107 (8.4%) | 164 (7.5%) | 2078 (6.4%) | 0.002 |
| **Z23** | 3245 (8.8%) | 75 (9.0%) | 100 (7.8%) | 197 (9.0%) | 2873 (8.8%) | 0.632 |
| **Z68** | 1934 (5.3%) | 37 (4.5%) | 37 (2.9%) | 82 (3.8%) | 1778 (5.5%) | <0.001 |
| **Z79** | 3746 (10.2%) | 181 (21.8%) | 243 (19.0%) | 343 (15.7%) | 2979 (9.2%) | <0.001 |
| **Z86** | 3026 (8.2%) | 93 (11.2%) | 118 (9.2%) | 234 (10.7%) | 2581 (7.9%) | <0.001 |
| **Z94** | 255 (0.7%) | 8 (1.0%) | 14 (1.1%) | 19 (0.9%) | 214 (0.7%) | 0.140 |
| **Z95** | 5246 (14.3%) | 357 (43.0%) | 479 (37.5%) | 610 (28.0%) | 3800 (11.7%) | <0.001 |
| **Z98** | 2429 (6.6%) | 104 (12.5%) | 155 (12.1%) | 242 (11.1%) | 1928 (5.9%) | <0.001 |
| **ACETAMINOPHEN** | 4920 (13.4%) | 229 (27.6%) | 328 (25.7%) | 446 (20.5%) | 3917 (12.1%) | <0.001 |
| **ALBUMIN** | 892 (2.4%) | 55 (6.6%) | 65 (5.1%) | 106 (4.9%) | 666 (2.0%) | <0.001 |
| **ALBUTEROL** | 2124 (5.8%) | 89 (10.7%) | 121 (9.5%) | 185 (8.5%) | 1729 (5.3%) | <0.001 |
| **AMIODARONE** | 1259 (3.4%) | 120 (14.4%) | 141 (11.1%) | 182 (8.3%) | 816 (2.5%) | <0.001 |
| **AMLODIPINE** | 2835 (7.7%) | 37 (4.5%) | 79 (6.2%) | 154 (7.1%) | 2565 (7.9%) | <0.001 |
| **APIXABAN** | 2847 (7.7%) | 157 (18.9%) | 206 (16.1%) | 326 (15.0%) | 2158 (6.6%) | <0.001 |
| **ASPIRIN** | 4165 (11.3%) | 228 (27.4%) | 342 (26.8%) | 412 (18.9%) | 3183 (9.8%) | <0.001 |
| **ATORVASTATIN** | 4381 (11.9%) | 215 (25.9%) | 302 (23.7%) | 433 (19.9%) | 3431 (10.6%) | <0.001 |
| **BUPIVACAINE** | 1635 (4.4%) | 76 (9.1%) | 109 (8.5%) | 143 (6.6%) | 1307 (4.0%) | <0.001 |
| **CALCIUM** | 1562 (4.2%) | 90 (10.8%) | 119 (9.3%) | 165 (7.6%) | 1188 (3.7%) | <0.001 |
| **CARVEDILOL** | 1722 (4.7%) | 160 (19.3%) | 209 (16.4%) | 232 (10.6%) | 1121 (3.4%) | <0.001 |
| **CEFAZOLIN** | 2569 (7.0%) | 128 (15.4%) | 188 (14.7%) | 238 (10.9%) | 2015 (6.2%) | <0.001 |
| **DEXAMETHASONE** | 1679 (4.6%) | 34 (4.1%) | 84 (6.6%) | 133 (6.1%) | 1428 (4.4%) | <0.001 |
| **DEXTROSE** | 2359 (6.4%) | 155 (18.7%) | 209 (16.4%) | 263 (12.1%) | 1732 (5.3%) | <0.001 |
| **DILTIAZEM** | 1205 (3.3%) | 41 (4.9%) | 66 (5.2%) | 112 (5.1%) | 986 (3.0%) | <0.001 |
| **DIPHENHYDRAMINE** | 2425 (6.6%) | 105 (12.6%) | 159 (12.5%) | 205 (9.4%) | 1956 (6.0%) | <0.001 |
| **FAMOTIDINE** | 2249 (6.1%) | 70 (8.4%) | 112 (8.8%) | 162 (7.4%) | 1905 (5.9%) | <0.001 |
| **FENTANYL** | 4075 (11.1%) | 211 (25.4%) | 311 (24.4%) | 403 (18.5%) | 3150 (9.7%) | <0.001 |
| **FUROSEMIDE** | 4059 (11.0%) | 310 (37.3%) | 396 (31.0%) | 489 (22.4%) | 2864 (8.8%) | <0.001 |
| **GABAPENTIN** | 1425 (3.9%) | 66 (7.9%) | 65 (5.1%) | 124 (5.7%) | 1170 (3.6%) | <0.001 |
| **GLUCAGON** | 1945 (5.3%) | 128 (15.4%) | 166 (13.0%) | 226 (10.4%) | 1425 (4.4%) | <0.001 |
| **GLUCOSE** | 2073 (5.6%) | 127 (15.3%) | 174 (13.6%) | 237 (10.9%) | 1535 (4.7%) | <0.001 |
| **HEPARIN** | 3773 (10.3%) | 215 (25.9%) | 306 (24.0%) | 382 (17.5%) | 2870 (8.8%) | <0.001 |
| **HYDRALAZINE** | 1232 (3.3%) | 58 (7.0%) | 73 (5.7%) | 106 (4.9%) | 995 (3.1%) | <0.001 |
| **HYDROMORPHONE** | 2036 (5.5%) | 89 (10.7%) | 135 (10.6%) | 187 (8.6%) | 1625 (5.0%) | <0.001 |
| **INSULIN** | 2214 (6.0%) | 144 (17.3%) | 179 (14.0%) | 242 (11.1%) | 1649 (5.1%) | <0.001 |
| **IOHEXOL** | 4709 (12.8%) | 205 (24.7%) | 286 (22.4%) | 383 (17.6%) | 3835 (11.8%) | <0.001 |
| **LACTATED** | 3299 (9.0%) | 133 (16.0%) | 213 (16.7%) | 296 (13.6%) | 2657 (8.2%) | <0.001 |
| **LIDOCAINE** | 5691 (15.5%) | 255 (30.7%) | 386 (30.3%) | 507 (23.3%) | 4543 (14.0%) | <0.001 |
| **LISINOPRIL** | 2079 (5.7%) | 74 (8.9%) | 133 (10.4%) | 187 (8.6%) | 1685 (5.2%) | <0.001 |
| **LORAZEPAM** | 1258 (3.4%) | 54 (6.5%) | 86 (6.7%) | 91 (4.2%) | 1027 (3.2%) | <0.001 |
| **LOSARTAN** | 2253 (6.1%) | 107 (12.9%) | 151 (11.8%) | 230 (10.6%) | 1765 (5.4%) | <0.001 |
| **MAGNESIUM** | 2820 (7.7%) | 167 (20.1%) | 224 (17.6%) | 295 (13.5%) | 2134 (6.6%) | <0.001 |
| **METOPROLOL** | 5772 (15.7%) | 303 (36.5%) | 418 (32.8%) | 620 (28.4%) | 4431 (13.6%) | <0.001 |
| **MIDAZOLAM** | 3581 (9.7%) | 193 (23.2%) | 282 (22.1%) | 362 (16.6%) | 2744 (8.4%) | <0.001 |
| **NALOXONE** | 2477 (6.7%) | 141 (17.0%) | 203 (15.9%) | 265 (12.2%) | 1868 (5.7%) | <0.001 |
| **NITROGLYCERIN** | 2557 (7.0%) | 141 (17.0%) | 230 (18.0%) | 277 (12.7%) | 1909 (5.9%) | <0.001 |
| **ONDANSETRON** | 4891 (13.3%) | 221 (26.6%) | 301 (23.6%) | 429 (19.7%) | 3940 (12.1%) | <0.001 |
| **OXYCODONE** | 1431 (3.9%) | 41 (4.9%) | 79 (6.2%) | 136 (6.2%) | 1175 (3.6%) | <0.001 |
| **PANTOPRAZOLE** | 3077 (8.4%) | 137 (16.5%) | 205 (16.1%) | 285 (13.1%) | 2450 (7.5%) | <0.001 |
| **PHENYLEPHRINE** | 2061 (5.6%) | 115 (13.8%) | 173 (13.6%) | 217 (10.0%) | 1556 (4.8%) | <0.001 |
| **POTASSIUM** | 2614 (7.1%) | 176 (21.2%) | 222 (17.4%) | 274 (12.6%) | 1942 (6.0%) | <0.001 |
| **PREDNISONE** | 1580 (4.3%) | 55 (6.6%) | 62 (4.9%) | 115 (5.3%) | 1348 (4.1%) | <0.001 |
| **PROPOFOL** | 3710 (10.1%) | 162 (19.5%) | 252 (19.7%) | 346 (15.9%) | 2950 (9.1%) | <0.001 |
| **ROCURONIUM** | 1779 (4.8%) | 87 (10.5%) | 126 (9.9%) | 202 (9.3%) | 1364 (4.2%) | <0.001 |
| **SODIUM** | 6931 (18.8%) | 327 (39.4%) | 449 (35.2%) | 571 (26.2%) | 5584 (17.2%) | <0.001 |
| **SPIRONOLACTONE** | 1399 (3.8%) | 198 (23.8%) | 232 (18.2%) | 217 (10.0%) | 752 (2.3%) | <0.001 |
| **VANCOMYCIN** | 1623 (4.4%) | 98 (11.8%) | 144 (11.3%) | 172 (7.9%) | 1209 (3.7%) | <0.001 |
| **VERAPAMIL** | 1976 (5.4%) | 118 (14.2%) | 168 (13.2%) | 208 (9.5%) | 1482 (4.6%) | <0.001 |
| **WARFARIN** | 848 (2.3%) | 86 (10.3%) | 95 (7.4%) | 100 (4.6%) | 567 (1.7%) | <0.001 |
| **Age** | 66.9 ± 16.3 (Median: 69.0, IQR: 58.0-79.0) | 68.8 ± 15.0 (Median: 71.0, IQR: 60.0-80.0) | 72.1 ± 12.2 (Median: 74.0, IQR: 64.0-81.0) | 71.4 ± 13.1 (Median: 73.0, IQR: 63.0-81.0) | 66.3 ± 16.5 (Median: 69.0, IQR: 57.0-78.0) | <0.001 |
| **Body Mass Index** | 24.6 ± 80.8 (Median: 26.8, IQR: 19.4-32.0) | 19.6 ± 14.9 (Median: 24.9, IQR: 0.0-29.8) | 22.9 ± 86.5 (Median: 25.2, IQR: 0.0-30.3) | 21.7 ± 14.3 (Median: 26.2, IQR: 0.0-31.5) | 25.0 ± 84.1 (Median: 26.9, IQR: 20.0-32.2) | <0.001 |
| **Diastolic Blood Pressure** | 59.8 ± 32.4 (Median: 72.0, IQR: 60.0-80.0) | 49.2 ± 36.1 (Median: 66.0, IQR: 0.0-78.0) | 52.9 ± 34.5 (Median: 68.0, IQR: 0.0-80.0) | 55.0 ± 33.8 (Median: 70.0, IQR: 0.0-80.0) | 60.7 ± 32.0 (Median: 72.0, IQR: 60.0-80.0) | <0.001 |
| **Systolic Blood Pressure** | 101.3 ± 54.6 (Median: 122.0, IQR: 102.0-135.0) | 82.0 ± 58.4 (Median: 110.0, IQR: 0.0-124.0) | 89.3 ± 57.7 (Median: 116.0, IQR: 0.0-130.0) | 93.5 ± 57.3 (Median: 118.0, IQR: 0.0-131.0) | 102.8 ± 54.0 (Median: 122.0, IQR: 105.0-136.0) | <0.001 |
| **Body Temperature (In Fahrenheit)** | 32.9 ± 46.1 (Median: 0.0, IQR: 0.0-97.2) | 28.4 ± 44.4 (Median: 0.0, IQR: 0.0-97.0) | 28.9 ± 44.5 (Median: 0.0, IQR: 0.0-97.0) | 31.5 ± 45.6 (Median: 0.0, IQR: 0.0-97.2) | 33.2 ± 46.3 (Median: 0.0, IQR: 0.0-97.2) | <0.001 |
| **Pulse rate** | 57.8 ± 34.1 (Median: 68.0, IQR: 52.0-80.0) | 53.4 ± 40.0 (Median: 67.0, IQR: 0.0-83.0) | 54.2 ± 37.7 (Median: 68.0, IQR: 0.0-80.0) | 55.8 ± 36.8 (Median: 67.0, IQR: 0.0-80.0) | 58.2 ± 33.5 (Median: 68.0, IQR: 53.0-80.0) | 0.046 |
| **White or Caucasian (First Race)** | 27880 (75.8%) | 606 (72.9%) | 1001 (78.4%) | 1765 (81.0%) | 24508 (75.4%) | <0.001 |
| **Black or African American (First Race)** | 3415 (9.3%) | 119 (14.3%) | 140 (11.0%) | 192 (8.8%) | 2964 (9.1%) | <0.001 |
| **Other (First Race)** | 3041 (8.3%) | 72 (8.7%) | 84 (6.6%) | 139 (6.4%) | 2746 (8.5%) | <0.001 |
| **Unknown (First Race)** | 1447 (3.9%) | 13 (1.6%) | 23 (1.8%) | 47 (2.2%) | 1364 (4.2%) | <0.001 |
| **Asian (First Race)** | 587 (1.6%) | 8 (1.0%) | 19 (1.5%) | 23 (1.1%) | 537 (1.7%) | <0.001 |
| **Patient Refused (First Race)** | 304 (0.8%) | 8 (1.0%) | 8 (0.6%) | 10 (0.5%) | 278 (0.9%) | <0.001 |
| **American Indian or Alaska Native (First Race)** | 73 (0.2%) | 4 (0.5%) | 1 (0.1%) | 2 (0.1%) | 66 (0.2%) | <0.001 |
| **Native Hawaiian or Other Pacific Islander (First Race)** | 37 (0.1%) | 1 (0.1%) | 0 (0.0%) | 2 (0.1%) | 34 (0.1%) | <0.001 |
| **Never (Smoking Status)** | 19425 (52.8%) | 305 (36.7%) | 510 (40.0%) | 961 (44.1%) | 17649 (54.3%) | <0.001 |
| **Former (Smoking Status)** | 14461 (39.3%) | 427 (51.4%) | 640 (50.2%) | 1027 (47.1%) | 12367 (38.1%) | <0.001 |
| **Every Day (Smoking Status)** | 1938 (5.3%) | 58 (7.0%) | 70 (5.5%) | 131 (6.0%) | 1679 (5.2%) | <0.001 |
| **Some Days (Smoking Status)** | 666 (1.8%) | 24 (2.9%) | 40 (3.1%) | 43 (2.0%) | 559 (1.7%) | <0.001 |
| **Never Assessed (Smoking Status)** | 136 (0.4%) | 12 (1.4%) | 11 (0.9%) | 11 (0.5%) | 102 (0.3%) | <0.001 |
| **Light Smoker (Smoking Status)** | 60 (0.2%) | 2 (0.2%) | 2 (0.2%) | 5 (0.2%) | 51 (0.2%) | <0.001 |
| **Unknown (Smoking Status)** | 33 (0.1%) | 0 (0.0%) | 0 (0.0%) | 0 (0.0%) | 33 (0.1%) | <0.001 |
| **Heavy Smoker (Smoking Status)** | 32 (0.1%) | 2 (0.2%) | 2 (0.2%) | 1 (0.0%) | 27 (0.1%) | <0.001 |
| **Passive Smoke Exposure - Never Smoker (Smoking Status)** | 13 (0.0%) | 0 (0.0%) | 0 (0.0%) | 0 (0.0%) | 13 (0.0%) | <0.001 |
| **Current Status Unknown (Smoking Status)** | 4 (0.0%) | 0 (0.0%) | 0 (0.0%) | 0 (0.0%) | 4 (0.0%) | <0.001 |
| **Male (Sex)** | 19017 (51.7%) | 628 (75.6%) | 941 (73.7%) | 1529 (70.1%) | 15919 (49.0%) | <0.001 |
| **Female (Sex)** | 17755 (48.3%) | 203 (24.4%) | 335 (26.3%) | 650 (29.8%) | 16567 (51.0%) | <0.001 |

**Supplemental Table 2:** Summary Statistics of predictor features in our temporal validation cohort (non-ECG features)

| Characteristic | Overall | Severely Reduced LVEF | Moderately Reduced LVEF | Mildly Reduced LVEF | Normal LVEF | P-Value |
|---|---|---|---|---|---|---|
| **C50** | 412 (2.1%) | 5 (1.2%) | 10 (1.5%) | 25 (2.1%) | 372 (2.1%) | 0.428 |
| **D50** | 755 (3.8%) | 20 (4.8%) | 33 (5.0%) | 52 (4.3%) | 650 (3.7%) | 0.159 |
| **D64** | 1110 (5.6%) | 30 (7.1%) | 41 (6.2%) | 74 (6.2%) | 965 (5.5%) | 0.3 |
| **E03** | 1256 (6.3%) | 28 (6.7%) | 41 (6.2%) | 60 (5.0%) | 1127 (6.4%) | 0.296 |
| **E11** | 2676 (13.4%) | 79 (18.8%) | 130 (19.5%) | 217 (18.1%) | 2250 (12.7%) | < 0.001 |
| **E55** | 1541 (7.7%) | 27 (6.4%) | 29 (4.4%) | 78 (6.5%) | 1407 (8.0%) | 0.001 |
| **E66** | 2301 (11.5%) | 34 (8.1%) | 53 (8.0%) | 115 (9.6%) | 2099 (11.9%) | < 0.001 |
| **E78** | 10080 (50.5%) | 207 (49.3%) | 382 (57.4%) | 672 (56.0%) | 8819 (49.9%) | < 0.001 |
| **E87** | 1132 (5.7%) | 52 (12.4%) | 55 (8.3%) | 97 (8.1%) | 928 (5.2%) | < 0.001 |
| **F41** | 1821 (9.1%) | 25 (6.0%) | 56 (8.4%) | 84 (7.0%) | 1656 (9.4%) | 0.004 |
| **G47** | 2137 (10.7%) | 42 (10.0%) | 70 (10.5%) | 153 (12.8%) | 1872 (10.6%) | 0.122 |
| **G89** | 1392 (7.0%) | 26 (6.2%) | 46 (6.9%) | 77 (6.4%) | 1243 (7.0%) | 0.789 |
| **I10** | 10887 (54.5%) | 219 (52.1%) | 382 (57.4%) | 712 (59.4%) | 9574 (54.1%) | 0.001 |
| **I21** | 613 (3.1%) | 41 (9.8%) | 59 (8.9%) | 112 (9.3%) | 401 (2.3%) | < 0.001 |
| **I25** | 4869 (24.4%) | 192 (45.7%) | 328 (49.2%) | 516 (43.0%) | 3833 (21.7%) | < 0.001 |
| **I34** | 1613 (8.1%) | 54 (12.9%) | 97 (14.6%) | 146 (12.2%) | 1316 (7.4%) | < 0.001 |
| **I35** | 2470 (12.4%) | 38 (9.0%) | 81 (12.2%) | 154 (12.8%) | 2197 (12.4%) | 0.203 |
| **I47** | 1267 (6.3%) | 75 (17.9%) | 97 (14.6%) | 145 (12.1%) | 950 (5.4%) | < 0.001 |
| **I48** | 5333 (26.7%) | 180 (42.9%) | 315 (47.3%) | 540 (45.0%) | 4298 (24.3%) | < 0.001 |
| **I49** | 1662 (8.3%) | 62 (14.8%) | 91 (13.7%) | 165 (13.8%) | 1344 (7.6%) | < 0.001 |
| **J44** | 817 (4.1%) | 24 (5.7%) | 41 (6.2%) | 68 (5.7%) | 684 (3.9%) | < 0.001 |
| **J45** | 907 (4.5%) | 13 (3.1%) | 21 (3.2%) | 42 (3.5%) | 831 (4.7%) | 0.029 |
| **K21** | 2598 (13.0%) | 53 (12.6%) | 85 (12.8%) | 162 (13.5%) | 2298 (13.0%) | 0.948 |
| **M25** | 1793 (9.0%) | 36 (8.6%) | 41 (6.2%) | 92 (7.7%) | 1624 (9.2%) | 0.019 |
| **M54** | 1991 (10.0%) | 37 (8.8%) | 50 (7.5%) | 92 (7.7%) | 1812 (10.2%) | 0.003 |
| **M79** | 1403 (7.0%) | 22 (5.2%) | 45 (6.8%) | 73 (6.1%) | 1263 (7.1%) | 0.249 |
| **N18** | 1783 (8.9%) | 64 (15.2%) | 109 (16.4%) | 147 (12.3%) | 1463 (8.3%) | < 0.001 |
| **N40** | 1100 (5.5%) | 30 (7.1%) | 59 (8.9%) | 81 (6.8%) | 930 (5.3%) | < 0.001 |
| **R00** | 2823 (14.1%) | 38 (9.0%) | 60 (9.0%) | 118 (9.8%) | 2607 (14.7%) | < 0.001 |
| **R06** | 4045 (20.3%) | 116 (27.6%) | 183 (27.5%) | 269 (22.4%) | 3477 (19.7%) | < 0.001 |
| **R07** | 2776 (13.9%) | 44 (10.5%) | 74 (11.1%) | 124 (10.3%) | 2534 (14.3%) | < 0.001 |
| **R10** | 1296 (6.5%) | 33 (7.9%) | 35 (5.3%) | 60 (5.0%) | 1168 (6.6%) | 0.052 |
| **R42** | 1384 (6.9%) | 19 (4.5%) | 26 (3.9%) | 66 (5.5%) | 1273 (7.2%) | < 0.001 |
| **R53** | 1275 (6.4%) | 24 (5.7%) | 38 (5.7%) | 82 (6.8%) | 1131 (6.4%) | 0.742 |
| **R60** | 1205 (6.0%) | 33 (7.9%) | 44 (6.6%) | 63 (5.3%) | 1065 (6.0%) | 0.247 |
| **R73** | 1612 (8.1%) | 24 (5.7%) | 50 (7.5%) | 69 (5.8%) | 1469 (8.3%) | 0.004 |
| **R94** | 2170 (10.9%) | 42 (10.0%) | 67 (10.1%) | 101 (8.4%) | 1960 (11.1%) | 0.029 |
| **Z00** | 3005 (15.1%) | 37 (8.8%) | 69 (10.4%) | 136 (11.3%) | 2763 (15.6%) | < 0.001 |
| **Z01** | 2564 (12.8%) | 52 (12.4%) | 91 (13.7%) | 160 (13.3%) | 2261 (12.8%) | 0.85 |
| **Z11** | 514 (2.6%) | 5 (1.2%) | 12 (1.8%) | 17 (1.4%) | 480 (2.7%) | 0.006 |
| **Z12** | 2928 (14.7%) | 41 (9.8%) | 75 (11.3%) | 143 (11.9%) | 2669 (15.1%) | < 0.001 |
| **Z17** | 240 (1.2%) | 3 (0.7%) | 4 (0.6%) | 16 (1.3%) | 217 (1.2%) | 0.37 |
| **Z20** | 197 (1.0%) | 4 (1.0%) | 11 (1.7%) | 11 (0.9%) | 171 (1.0%) | 0.369 |
| **Z23** | 1136 (5.7%) | 18 (4.3%) | 44 (6.6%) | 65 (5.4%) | 1009 (5.7%) | 0.431 |
| **Z68** | 1314 (6.6%) | 12 (2.9%) | 28 (4.2%) | 59 (4.9%) | 1215 (6.9%) | < 0.001 |
| **Z79** | 2207 (11.1%) | 83 (19.8%) | 106 (15.9%) | 200 (16.7%) | 1818 (10.3%) | < 0.001 |
| **Z86** | 1876 (9.4%) | 42 (10.0%) | 64 (9.6%) | 139 (11.6%) | 1631 (9.2%) | 0.054 |
| **Z94** | 165 (0.8%) | 1 (0.2%) | 1 (0.2%) | 6 (0.5%) | 157 (0.9%) | 0.049 |
| **Z95** | 3151 (15.8%) | 167 (39.8%) | 239 (35.9%) | 378 (31.5%) | 2367 (13.4%) | < 0.001 |
| **Z98** | 1597 (8.0%) | 56 (13.3%) | 83 (12.5%) | 132 (11.0%) | 1326 (7.5%) | < 0.001 |
| **ACETAMINOPHEN** | 3280 (16.4%) | 131 (31.2%) | 186 (27.9%) | 309 (25.8%) | 2654 (15.0%) | < 0.001 |
| **ALBUMIN** | 626 (3.1%) | 30 (7.1%) | 35 (5.3%) | 75 (6.3%) | 486 (2.7%) | < 0.001 |
| **ALBUTEROL** | 1479 (7.4%) | 47 (11.2%) | 72 (10.8%) | 111 (9.3%) | 1249 (7.1%) | < 0.001 |
| **AMIODARONE** | 797 (4.0%) | 65 (15.5%) | 86 (12.9%) | 114 (9.5%) | 532 (3.0%) | < 0.001 |
| **AMLODIPINE** | 1789 (9.0%) | 21 (5.0%) | 46 (6.9%) | 102 (8.5%) | 1620 (9.2%) | 0.005 |
| **APIXABAN** | 2232 (11.2%) | 110 (26.2%) | 164 (24.6%) | 225 (18.8%) | 1733 (9.8%) | < 0.001 |
| **ASPIRIN** | 2741 (13.7%) | 120 (28.6%) | 172 (25.8%) | 291 (24.3%) | 2158 (12.2%) | < 0.001 |
| **ATORVASTATIN** | 2679 (13.4%) | 92 (21.9%) | 164 (24.6%) | 233 (19.4%) | 2190 (12.4%) | < 0.001 |
| **BUPIVACAINE** | 1039 (5.2%) | 31 (7.4%) | 54 (8.1%) | 101 (8.4%) | 853 (4.8%) | < 0.001 |
| **CALCIUM** | 1091 (5.5%) | 57 (13.6%) | 70 (10.5%) | 127 (10.6%) | 837 (4.7%) | < 0.001 |
| **CARVEDILOL** | 977 (4.9%) | 61 (14.5%) | 110 (16.5%) | 136 (11.3%) | 670 (3.8%) | < 0.001 |
| **CEFAZOLIN** | 1800 (9.0%) | 64 (15.2%) | 97 (14.6%) | 173 (14.4%) | 1466 (8.3%) | < 0.001 |
| **DEXAMETHASONE** | 1166 (5.8%) | 24 (5.7%) | 47 (7.1%) | 84 (7.0%) | 1011 (5.7%) | 0.154 |
| **DEXTROSE** | 1527 (7.6%) | 79 (18.8%) | 102 (15.3%) | 180 (15.0%) | 1166 (6.6%) | < 0.001 |
| **DILTIAZEM** | 732 (3.7%) | 20 (4.8%) | 39 (5.9%) | 70 (5.8%) | 603 (3.4%) | < 0.001 |
| **DIPHENHYDRAMINE** | 1425 (7.1%) | 50 (11.9%) | 67 (10.1%) | 134 (11.2%) | 1174 (6.6%) | < 0.001 |
| **FAMOTIDINE** | 1061 (5.3%) | 26 (6.2%) | 43 (6.5%) | 80 (6.7%) | 912 (5.2%) | 0.055 |
| **FENTANYL** | 2697 (13.5%) | 115 (27.4%) | 165 (24.8%) | 286 (23.9%) | 2131 (12.1%) | < 0.001 |
| **FUROSEMIDE** | 2486 (12.5%) | 176 (41.9%) | 223 (33.5%) | 293 (24.4%) | 1794 (10.1%) | < 0.001 |
| **GABAPENTIN** | 862 (4.3%) | 20 (4.8%) | 42 (6.3%) | 70 (5.8%) | 730 (4.1%) | 0.002 |
| **GLUCAGON** | 1293 (6.5%) | 69 (16.4%) | 88 (13.2%) | 161 (13.4%) | 975 (5.5%) | < 0.001 |
| **GLUCOSE** | 1346 (6.7%) | 68 (16.2%) | 92 (13.8%) | 164 (13.7%) | 1022 (5.8%) | < 0.001 |
| **HEPARIN** | 2361 (11.8%) | 121 (28.8%) | 157 (23.6%) | 252 (21.0%) | 1831 (10.4%) | < 0.001 |
| **HYDRALAZINE** | 922 (4.6%) | 21 (5.0%) | 42 (6.3%) | 72 (6.0%) | 787 (4.5%) | 0.013 |
| **HYDROMORPHONE** | 1508 (7.6%) | 53 (12.6%) | 78 (11.7%) | 138 (11.5%) | 1239 (7.0%) | < 0.001 |
| **INSULIN** | 1462 (7.3%) | 77 (18.3%) | 97 (14.6%) | 169 (14.1%) | 1119 (6.3%) | < 0.001 |
| **IOHEXOL** | 3295 (16.5%) | 112 (26.7%) | 162 (24.3%) | 297 (24.8%) | 2724 (15.4%) | < 0.001 |
| **LACTATED** | 2227 (11.2%) | 80 (19.0%) | 115 (17.3%) | 196 (16.3%) | 1836 (10.4%) | < 0.001 |
| **LIDOCAINE** | 3749 (18.8%) | 139 (33.1%) | 209 (31.4%) | 346 (28.9%) | 3055 (17.3%) | < 0.001 |
| **LISINOPRIL** | 988 (4.9%) | 27 (6.4%) | 45 (6.8%) | 86 (7.2%) | 830 (4.7%) | < 0.001 |
| **LORAZEPAM** | 661 (3.3%) | 29 (6.9%) | 34 (5.1%) | 52 (4.3%) | 546 (3.1%) | < 0.001 |
| **LOSARTAN** | 1596 (8.0%) | 57 (13.6%) | 86 (12.9%) | 160 (13.3%) | 1293 (7.3%) | < 0.001 |
| **MAGNESIUM** | 1807 (9.1%) | 98 (23.3%) | 114 (17.1%) | 212 (17.7%) | 1383 (7.8%) | < 0.001 |
| **METOPROLOL** | 3646 (18.3%) | 178 (42.4%) | 230 (34.5%) | 420 (35.0%) | 2818 (15.9%) | < 0.001 |
| **MIDAZOLAM** | 2290 (11.5%) | 100 (23.8%) | 138 (20.7%) | 247 (20.6%) | 1805 (10.2%) | < 0.001 |
| **NALOXONE** | 1705 (8.5%) | 75 (17.9%) | 110 (16.5%) | 179 (14.9%) | 1341 (7.6%) | < 0.001 |
| **NITROGLYCERIN** | 1697 (8.5%) | 81 (19.3%) | 116 (17.4%) | 206 (17.2%) | 1294 (7.3%) | < 0.001 |
| **ONDANSETRON** | 3383 (16.9%) | 117 (27.9%) | 170 (25.5%) | 320 (26.7%) | 2776 (15.7%) | < 0.001 |
| **OXYCODONE** | 1260 (6.3%) | 41 (9.8%) | 60 (9.0%) | 111 (9.3%) | 1048 (5.9%) | < 0.001 |
| **PANTOPRAZOLE** | 2057 (10.3%) | 85 (20.2%) | 107 (16.1%) | 191 (15.9%) | 1674 (9.5%) | < 0.001 |
| **PHENYLEPHRINE** | 1402 (7.0%) | 62 (14.8%) | 84 (12.6%) | 154 (12.8%) | 1102 (6.2%) | < 0.001 |
| **POTASSIUM** | 1729 (8.7%) | 106 (25.2%) | 122 (18.3%) | 202 (16.8%) | 1299 (7.3%) | < 0.001 |
| **PREDNISONE** | 1208 (6.1%) | 29 (6.9%) | 45 (6.8%) | 73 (6.1%) | 1061 (6.0%) | 0.752 |
| **PROPOFOL** | 2438 (12.2%) | 89 (21.2%) | 132 (19.8%) | 233 (19.4%) | 1984 (11.2%) | < 0.001 |
| **ROCURONIUM** | 1254 (6.3%) | 46 (11.0%) | 65 (9.8%) | 124 (10.3%) | 1019 (5.8%) | < 0.001 |
| **SODIUM** | 4408 (22.1%) | 161 (38.3%) | 236 (35.4%) | 406 (33.9%) | 3605 (20.4%) | < 0.001 |
| **SPIRONOLACTONE** | 1037 (5.2%) | 119 (28.3%) | 120 (18.0%) | 182 (15.2%) | 616 (3.5%) | < 0.001 |
| **VANCOMYCIN** | 1084 (5.4%) | 46 (11.0%) | 66 (9.9%) | 118 (9.8%) | 854 (4.8%) | < 0.001 |
| **VERAPAMIL** | 1235 (6.2%) | 56 (13.3%) | 90 (13.5%) | 144 (12.0%) | 945 (5.3%) | < 0.001 |
| **WARFARIN** | 292 (1.5%) | 27 (6.4%) | 22 (3.3%) | 41 (3.4%) | 202 (1.1%) | < 0.001 |
| **Age** | 67.7 ± 15.6 (Median: 70.0, IQR: 60.0-79.0) | 70.3 ± 12.6 (Median: 71.0, IQR: 63.0-79.0) | 72.6 ± 12.4 (Median: 74.0, IQR: 65.0-81.0) | 70.3 ± 13.0 (Median: 71.0, IQR: 62.0-79.0) | 67.2 ± 15.8 (Median: 70.0, IQR: 59.0-78.0) | < 0.001 |
| **Body Mass Index** | 24.1 ± 94.1 (Median: 26.9, IQR: 20.2-32.0) | 20.8 ± 14.0 (Median: 25.1, IQR: 0.0-29.5) | 21.4 ± 13.7 (Median: 25.4, IQR: 0.0-30.5) | 22.3 ± 14.6 (Median: 26.3, IQR: 0.0-31.6) | 24.4 ± 99.9 (Median: 27.1, IQR: 20.6-32.2) | < 0.001 |
| **Diastolic Blood Pressure** | 59.4 ± 31.8 (Median: 70.0, IQR: 60.0-80.0) | 51.8 ± 33.3 (Median: 66.0, IQR: 0.0-77.0) | 55.1 ± 32.6 (Median: 68.0, IQR: 54.0-78.0) | 55.7 ± 33.1 (Median: 70.0, IQR: 50.0-80.0) | 60.0 ± 31.6 (Median: 71.0, IQR: 60.0-80.0) | < 0.001 |
| **Systolic Blood Pressure** | 101.5 ± 54.0 (Median: 122.0, IQR: 103.0-134.0) | 86.8 ± 53.3 (Median: 110.0, IQR: 0.0-122.0) | 93.2 ± 54.9 (Median: 114.0, IQR: 88.5-130.0) | 94.4 ± 55.7 (Median: 118.0, IQR: 89.0-130.0) | 102.7 ± 53.7 (Median: 122.0, IQR: 104.0-135.0) | < 0.001 |
| **Body Temperature (In Fahrenheit)** | 5.7 ± 22.9 (Median: 0.0, IQR: 0.0-0.0) | 6.5 ± 24.5 (Median: 0.0, IQR: 0.0-0.0) | 5.7 ± 23.0 (Median: 0.0, IQR: 0.0-0.0) | 5.0 ± 21.6 (Median: 0.0, IQR: 0.0-0.0) | 5.7 ± 22.9 (Median: 0.0, IQR: 0.0-0.0) | 0.705 |
| **Pulse rate** | 57.7 ± 33.4 (Median: 68.0, IQR: 52.0-79.0) | 59.5 ± 39.2 (Median: 73.0, IQR: 0.0-85.0) | 56.3 ± 36.1 (Median: 67.0, IQR: 0.0-79.0) | 55.4 ± 36.1 (Median: 66.0, IQR: 0.0-79.0) | 57.9 ± 32.9 (Median: 68.0, IQR: 53.0-79.0) | 0.002 |
| **White or Caucasian (First Race)** | 15519 (77.7%) | 326 (77.6%) | 527 (79.1%) | 966 (80.6%) | 13700 (77.5%) | 0.026 |
| **Black or African American (First Race)** | 1780 (8.9%) | 49 (11.7%) | 71 (10.7%) | 114 (9.5%) | 1546 (8.7%) | 0.026 |
| **Other (First Race)** | 1575 (7.9%) | 29 (6.9%) | 43 (6.5%) | 77 (6.4%) | 1426 (8.1%) | 0.026 |
| **Unknown (First Race)** | 481 (2.4%) | 7 (1.7%) | 6 (0.9%) | 17 (1.4%) | 451 (2.6%) | 0.026 |
| **Asian (First Race)** | 346 (1.7%) | 6 (1.4%) | 9 (1.4%) | 12 (1.0%) | 319 (1.8%) | 0.026 |
| **Patient Refused (First Race)** | 192 (1.0%) | 2 (0.5%) | 8 (1.2%) | 7 (0.6%) | 175 (1.0%) | 0.026 |
| **American Indian or Alaska Native (First Race)** | 48 (0.2%) | 0 (0.0%) | 1 (0.2%) | 4 (0.3%) | 43 (0.2%) | 0.026 |
| **Native Hawaiian or Other Pacific Islander (First Race)** | 24 (0.1%) | 1 (0.2%) | 1 (0.2%) | 2 (0.2%) | 20 (0.1%) | 0.026 |
| **Other/Unspecified (First Race)** | 1 (0.0%) | 0 (0.0%) | 0 (0.0%) | 0 (0.0%) | 1 (0.0%) | 0.026 |
| **Never (Smoking Status)** | 10626 (53.2%) | 174 (41.4%) | 278 (41.7%) | 528 (44.0%) | 9646 (54.6%) | <0.001 |
| **Former (Smoking Status)** | 7832 (39.2%) | 206 (49.0%) | 329 (49.4%) | 540 (45.0%) | 6757 (38.2%) | <0.001 |
| **Every Day (Smoking Status)** | 1010 (5.1%) | 24 (5.7%) | 38 (5.7%) | 95 (7.9%) | 853 (4.8%) | <0.001 |
| **Some Days (Smoking Status)** | 353 (1.8%) | 12 (2.9%) | 20 (3.0%) | 30 (2.5%) | 291 (1.6%) | <0.001 |
| **Never Assessed (Smoking Status)** | 47 (0.2%) | 2 (0.5%) | 0 (0.0%) | 1 (0.2%) | 44 (0.2%) | <0.001 |
| **Light Smoker (Smoking Status)** | 29 (0.1%) | 0 (0.0%) | 1 (0.2%) | 2 (0.2%) | 26 (0.1%) | <0.001 |
| **Unknown (Smoking Status)** | 51 (0.2%) | 1 (0.2%) | 0 (0.0%) | 1 (0.1%) | 49 (0.2%) | <0.001 |
| **Heavy Smoker (Smoking Status)** | 10 (0.1%) | 1 (0.2%) | 0 (0.0%) | 1 (0.1%) | 8 (0.0%) | <0.001 |
| **Passive Smoke Exposure - Never Smoker (Smoking Status)** | 7 (0.0%) | 0 (0.0%) | 0 (0.0%) | 0 (0.0%) | 7 (0.0%) | <0.001 |
| **Current Status Unknown (Smoking Status)** | 1 (0.0%) | 0 (0.0%) | 0 (0.0%) | 0 (0.0%) | 1 (0.0%) | <0.001 |
| **Male (Sex)** | 10468 (52.4%) | 321 (76.4%) | 486 (73.0%) | 863 (72.0%) | 8798 (49.8%) | <0.001 |
| **Female (Sex)** | 9493 (47.5%) | 99 (23.6%) | 180 (27.0%) | 336 (28.0%) | 8878 (50.2%) | <0.001 |
| **Unknown (Sex)** | 4 (0.0%) | 0 (0.0%) | 0 (0.0%) | 0 (0.0%) | 4 (0.0%) | <0.001 |
| **Other (Sex)** | 1 (0.0%) | 0 (0.0%) | 0 (0.0%) | 0 (0.0%) | 1 (0.0%) | <0.001 |

## Supplemental Note 1

**Interpretation of Figure 3 (in main manuscript): How to read the plot.** Each point corresponds to one ECG–EHR record. The x-axis is the SHAP value (feature contribution) for the Normal (≥50%) class output.

- Points *to the right (positive SHAP)* increase the model's "Normal LVEF" score (push toward "Normal").
- Points *to the left (negative SHAP)* decrease the "Normal LVEF" score (push away from "Normal," favoring reduced-LVEF classes).

  Color indicates feature magnitude: red = higher feature value, blue = lower feature value. For binary variables (e.g., Sex), red corresponds to Sex=1 (male) and blue corresponds to Sex=0 (female).

*Remark: These SHAP analyses identify features that the model relies on when assigning the Normal LVEF category. The direction and magnitude of SHAP contributions should be interpreted as associations within this cohort and feature representation, not as evidence of causality. Nonetheless, the convergence of high-impact ECG amplitude descriptors with clinically plausible EHR correlates (cardiomyopathy, ischemic cardiomyopathy, blood pressure, and sex) supports face validity of the model's learned decision rules.*

**1) High-impact clinical markers (history & vitals)**

Several EHR variables act as strong correlates of whether the model assigns a patient to Normal LVEF.

- Presence of cardiomyopathy (ICD-10 I42.9) history contributes away from "Normal LVEF" (negative SHAP), consistent with cardiomyopathy being a high-risk substrate for reduced systolic function.
- Ischemic cardiomyopathy (ICD-10 I25.5) similarly contributes away from "Normal LVEF", reflecting the known association between coronary disease with myocardial dysfunction.
- Nonrheumatic aortic stenosis (ICD-10 I35.0) tends to increase the model's Normal (≥50%) score (positive SHAP) in this cohort, likely related to many outpatient aortic stenosis evaluations occurring in patients with preserved EF. This should be interpreted as an association in this dataset rather than implying that aortic stenosis protects against reduced EF.
- Variations in systolic blood pressure contribute to the "Normal LVEF" class scoring; lower BP frequently aligns with reduced Normal score, consistent with advanced heart failure physiology (reduced forward output) in some patients.
- Sex is one of the strongest contributors for the this class. The red points (Male, Sex=1) show a different SHAP distribution compared to the blue points (Female, Sex=0), indicating that sex functions as a demographic correlate used by the model in this cohort. Again, this should be interpreted as a dataset-level association rather than a causal effect.

**2) ECG voltage summaries and rhythm-related morphology**

A large fraction of influential ECG predictors are voltage/amplitude summaries—especially in lateral or limb leads.

- **Lead I QR-interval amplitude statistics (average and median)**: These are among the top ECG predictors for the Normal score. The beeswarm suggests that certain ranges of Lead I QR amplitude systematically increase or decrease the "Normal LVEF" score, indicating that the model uses limb-lead depolarization amplitude as a marker of normal vs abnormal systolic function.
- **Lead aVR and V4 amplitude-band descriptors (lower-percentile negative/low amplitude bands)**: These features reflect how often the signal occupies low-amplitude regimes. Shifts in these distributions contribute to the "Normal LVEF" score, consistent with low voltage/attenuated morphology patterns in structural disease.

Beyond depolarization amplitude, the model also uses repolarization-related descriptors:

- **Lead V5 ST-segment voltage (average)** and **Lead V5 RS-interval voltage (median)**: These features contribute to the "Normal LVEF" score, consistent with lateral-lead repolarization/depolarization morphology being informative for myocardial dysfunction and ischemia-related changes.

**4) Signal complexity / frequency-domain features**

- **Spectral entropy (Lead V6)**: Frequency-domain complexity is among the top contributors for the Normal class. Higher entropy (a broader, more complex frequency distribution) tends to shift the "Normal LVEF" score in a consistent direction in this cohort. This likely reflects that certain abnormal morphologies yield more stereotyped waveform patterns, whereas normal rhythms may exhibit different complexity profiles;

however, this remains a model-derived signal correlate rather than a physiologic conclusion.

**Supplemental Figure S2.** Class-specific SHAP analysis for class “severely reduced” LVEF (below 30%), positive SHAP pushes toward severe reduction.

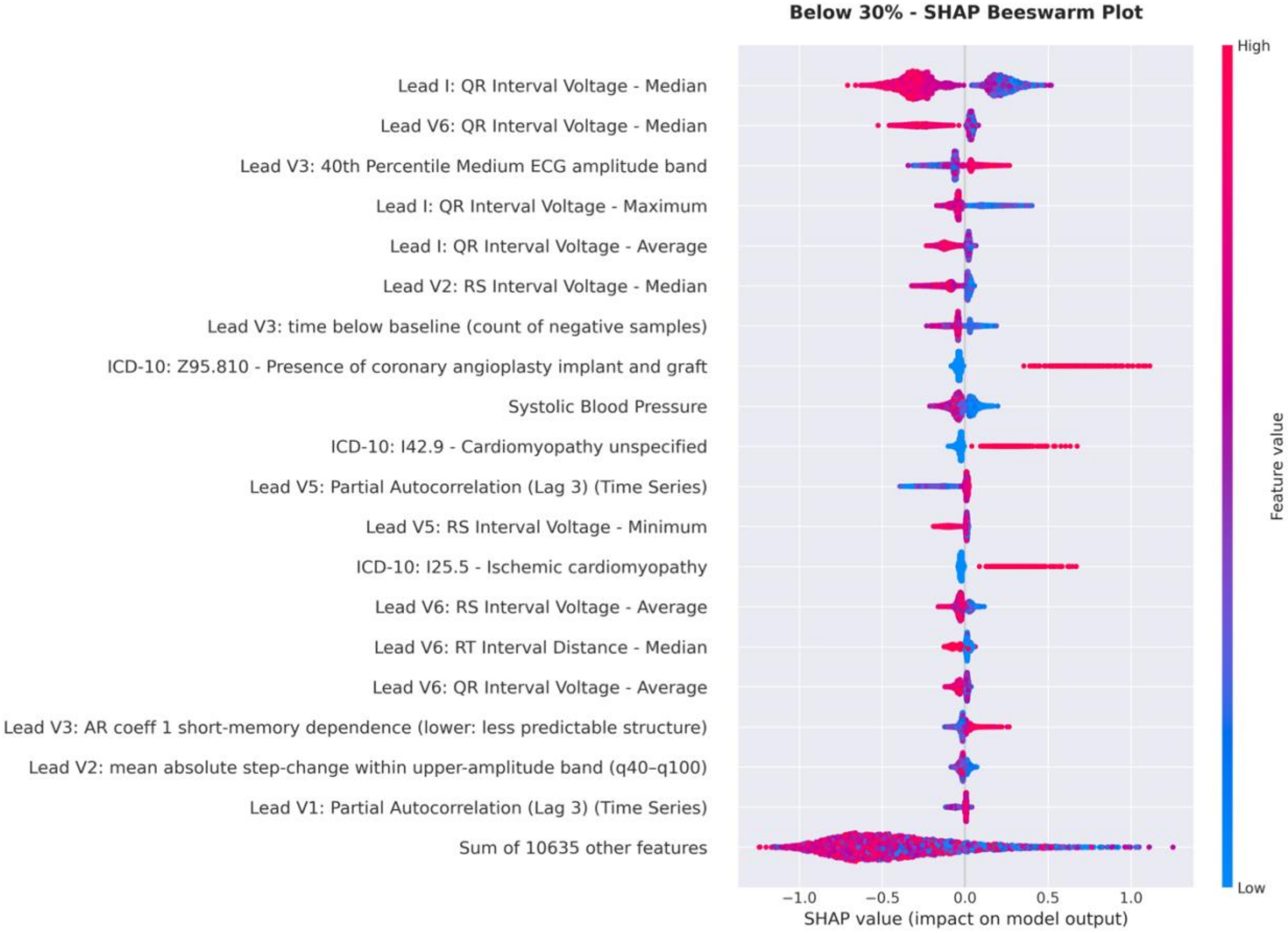

**Supplemental Figure S3.** Class-specific SHAP analysis for class “moderately reduced” LVEF (30-40%), positive SHAP pushes toward moderate reduction.

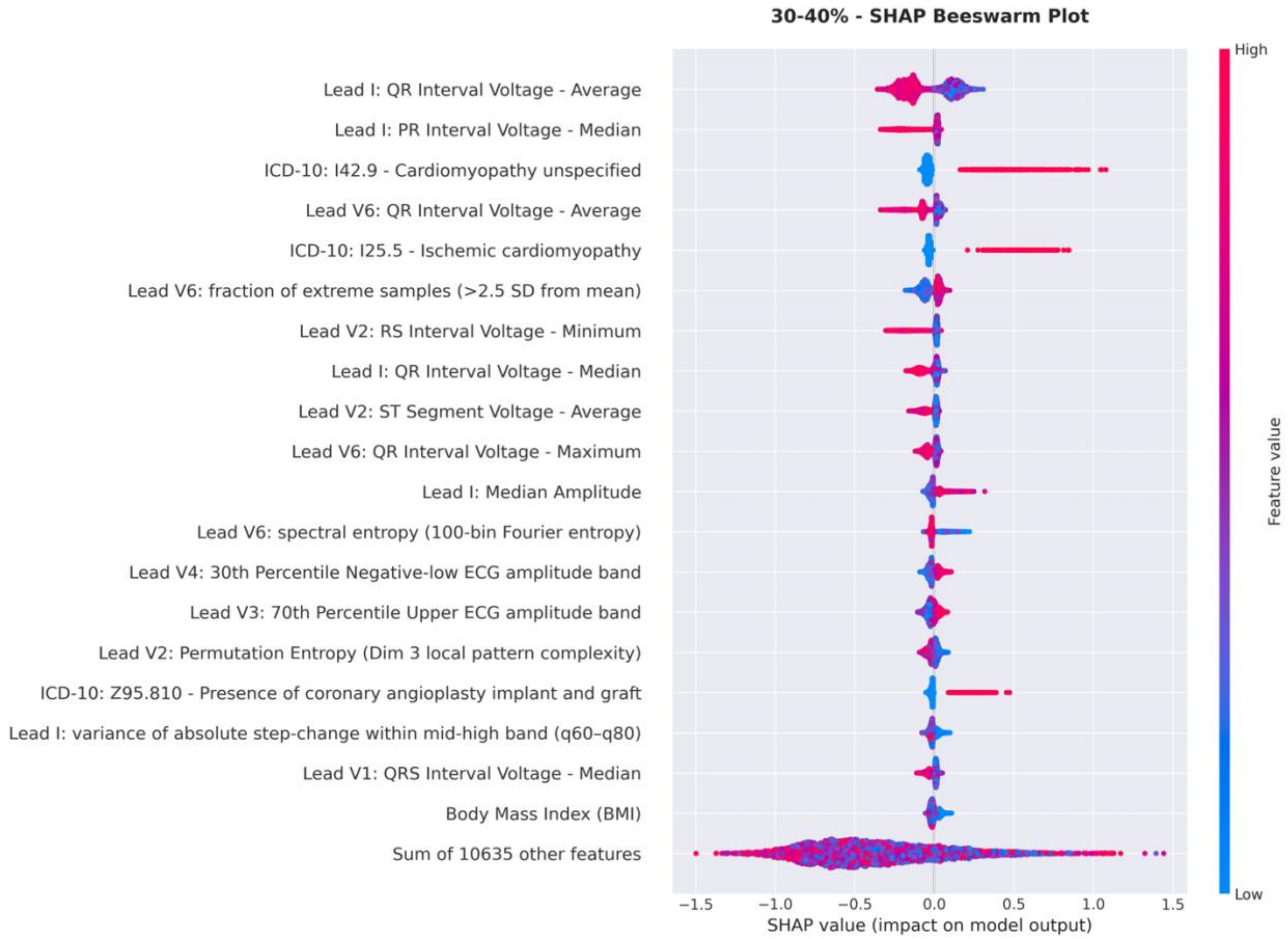

**Supplemental Figure S4.** Class-specific SHAP analysis for class "mildly reduced" LVEF (40-50%), positive SHAP pushes toward mild reduction.

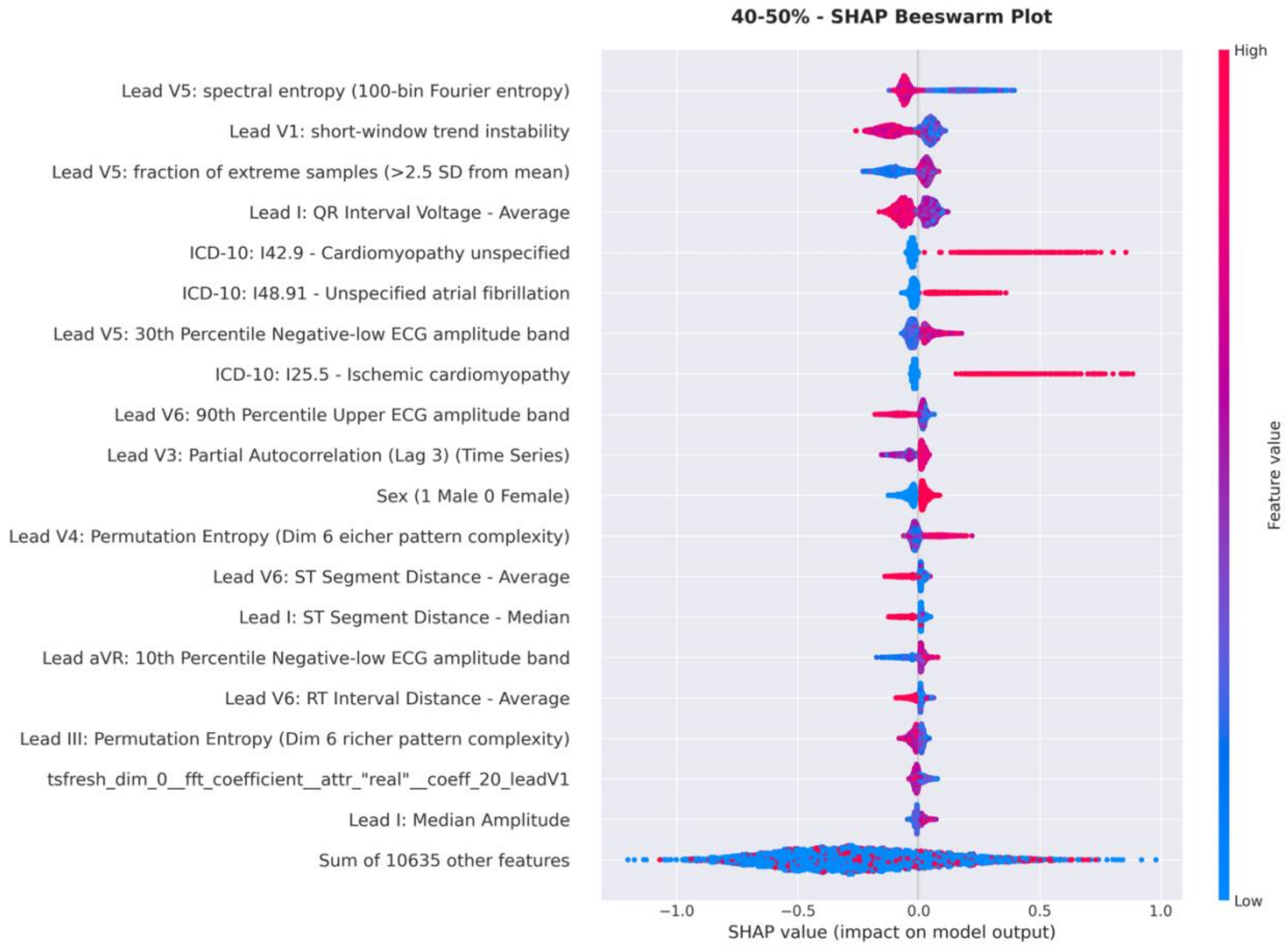

**Supplemental Figure S5.** Feature ranking stability plots under bootstrap resampling (B=20) of the held-out test set.

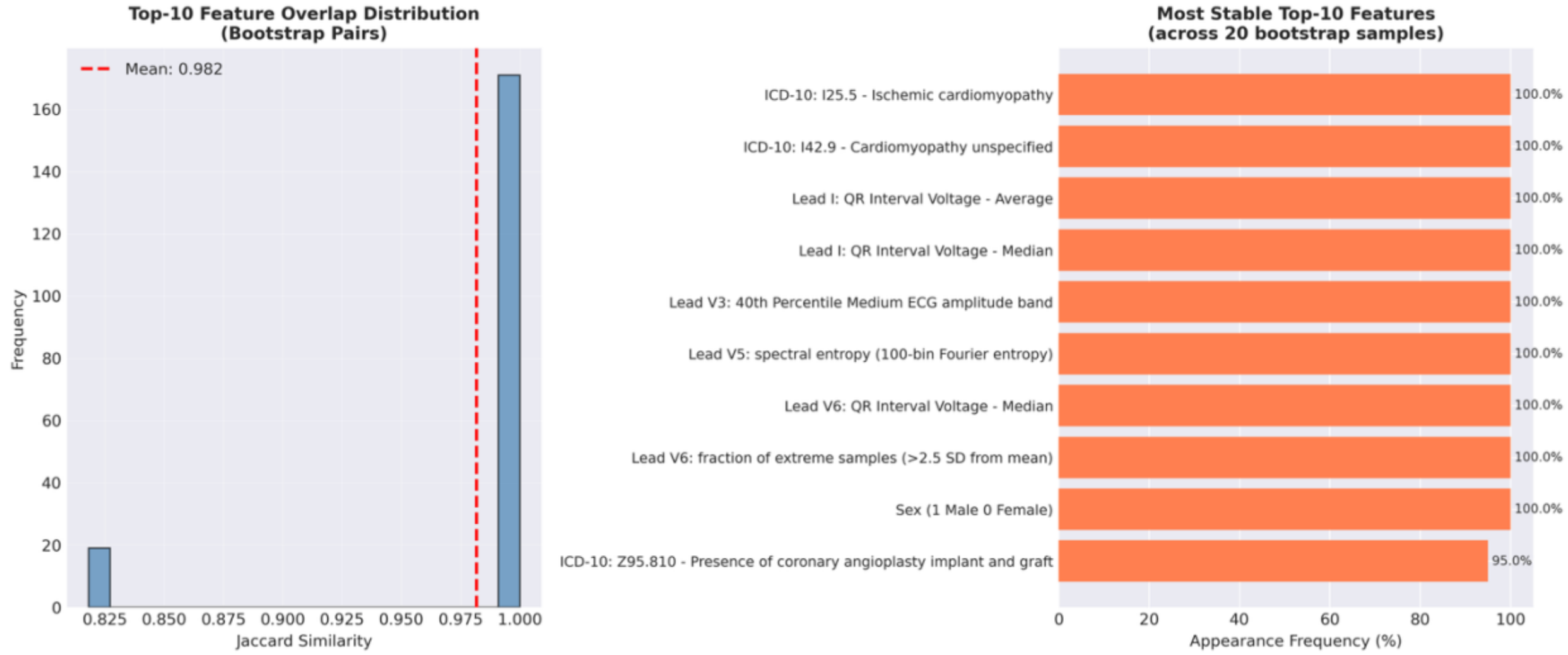

**Supplemental Figure S6.** SHAP dependence plots visualize the relationship between the top 2 continuous ECG-derived features and their contribution to the model output, highlighting directionality and potential non-linear effects learned by the classifier. (Voltage is the same as amplitude in ECG signals.)

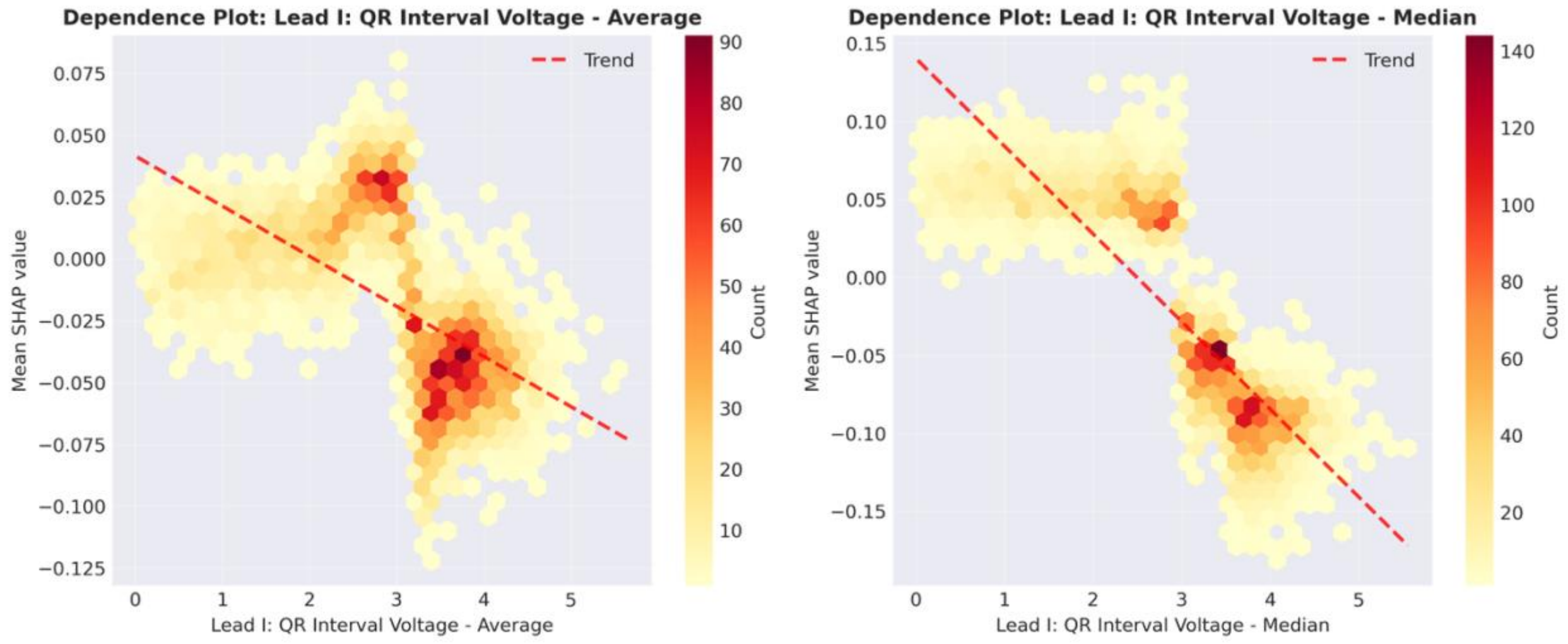

## Supplemental Figure S7.

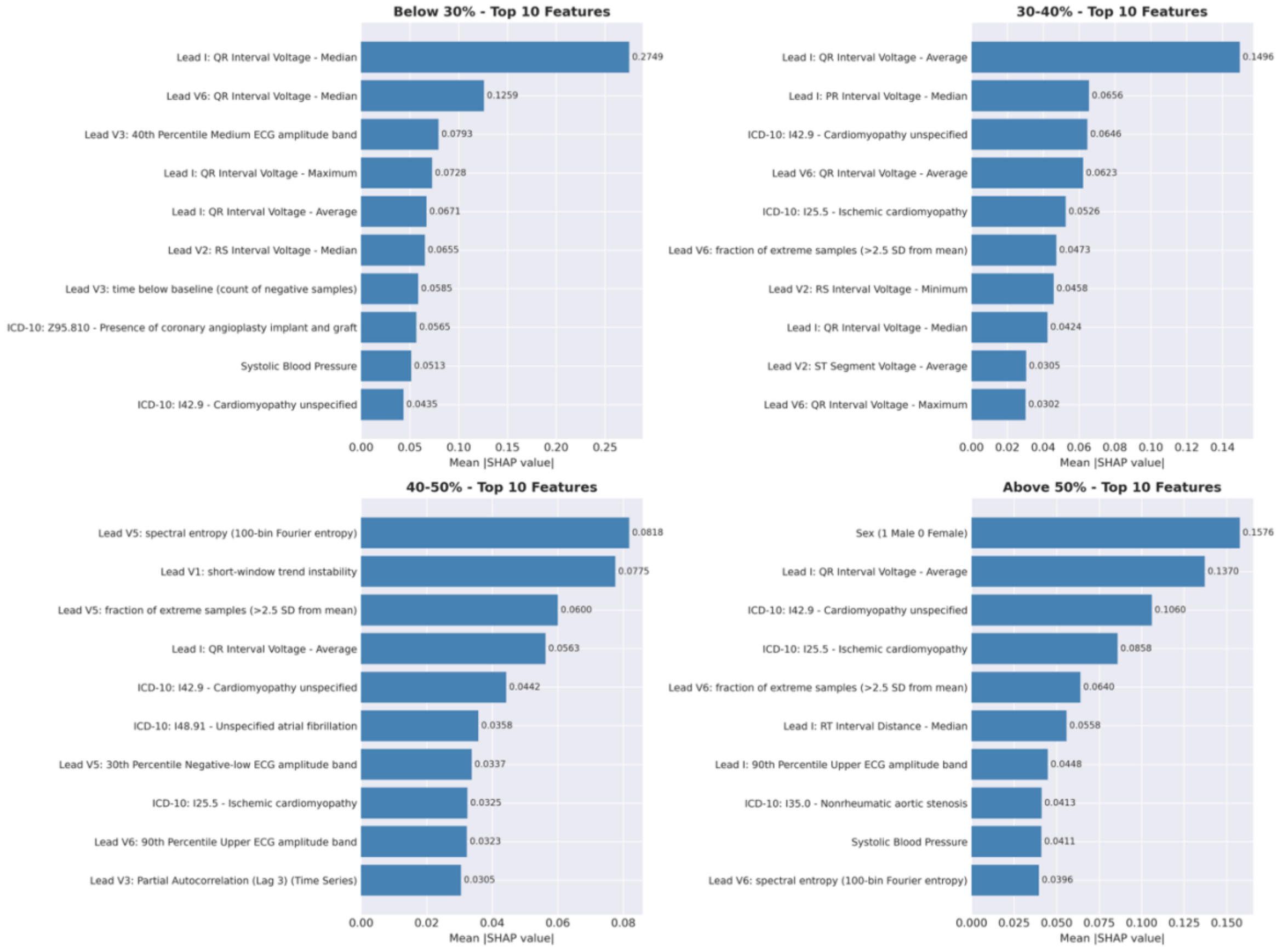

# Software Specifications

After querying the study or test cohort from the hospital database, we performed all input data preprocessing, model training and evaluation in **Python** (e.g., NeuroKit2's `ecg_clean()` function for noise removal, and scikit-learn's `preprocessing.scale()`for standardization etc.).**Our model was implemented with the `XGBClassifier` from the `xgboost` library with training parameters tuned to learning rate: 0.08, maximum depth: 7, minimum child weight: 5, evaluation metric: `mlogloss`, and objective: `multi:softprob` for multi-class classification.** These parameters were finetuned to achieve best per-class AUC scores while maintaining reasonable per-class sensitivity and specificity. **Training time without hyperparameter search was around 3.33 minutes on a single NVidia L40s GPU with 48GB RAM. Bootstrapping with ten random resamples of the internal hold-out and external temporal test sets to construct 95% confidence intervals was completed in 0.12 to 0.17 minutes.**